\documentclass[letterpaper]{article}
\usepackage[preprint]{aaai2027}
\usepackage[hyphens]{url}
\usepackage{graphicx}
\usepackage{natbib}
\usepackage{caption}
\usepackage{algorithm}
\usepackage{algorithmic}
\usepackage{newfloat}
\usepackage{listings}
\DeclareCaptionStyle{ruled}{labelfont=normalfont,labelsep=colon,strut=off}
\floatstyle{ruled}
\newfloat{listing}{tb}{lst}{}
\floatname{listing}{Listing}
\usepackage{booktabs}
\usepackage{colortbl}
\usepackage{tabularx}
\usepackage{makecell}
\usepackage{multirow}
\usepackage{rotating}

\begingroup
  \catcode`\_=12 \catcode`\~=12 \catcode`\#=12 \catcode`\&=12 \catcode`\%=12
  \gdef\ProjectPageURI{https://jigsaw0612.github.io/SparSTAR_project_page/}
\endgroup
\newcommand{\ProjectPageText}{{\urlstyle{tt}\url{https://jigsaw0612.github.io/SparSTAR_project_page/}}}
\newcommand{\ProjectPageLink}{%
  \leavevmode
  \pdfstartlink attr{/Border[0 0 0]}
    user{/Subtype/Link/A<</Type/Action/S/URI/URI(\ProjectPageURI)>>}%
  \ProjectPageText
  \pdfendlink
}

\title{SparSTAR: Sparse Attention for SpaceTime AutoRegressive Video Synthesis}
\author{
  Jongbeom Lee\textsuperscript{\rm 1*},
  Hyunwoo Yu\textsuperscript{\rm 1*},
  Jincheol Yang\textsuperscript{\rm 1},
  Jaemin Choi\textsuperscript{\rm 1},
  Suk-Ju Kang\textsuperscript{\rm 1\textdagger}
}
\affiliations{
  \textsuperscript{\rm 1}Sogang University\\
  Seoul, Republic of Korea\\
  *Equal contribution\qquad \textdagger Corresponding author\\
  Project page: \ProjectPageLink
}

\begin{document}
\maketitle

\begin{abstract}
InfinityStar extends visual autoregressive generation to video through a sequence of image and clip pyramids. Its changing scale and cross-clip context, however, leave late-scale attention costly and make sparse patterns reused from diffusion or image VAR models unreliable. We introduce SparSTAR, a training-free block-sparse attention method tailored to this setting. At each expensive scale and attention head, SparSTAR scores contiguous key blocks from the current query and key activations, retains required conditioning context, and executes the selected blocks through a forward-only sparse path. We analyze cross-scale consistency within a clip, pattern persistence across clip boundaries, and quality degradation as reuse spans increasingly distant scales. Across these analyses, important key blocks shift, showing that recomputing block selection at each target scale is more reliable than reusing a transferred mask. On 720p text-to-video and image-to-video generation, SparSTAR preserves every token and refinement scale while providing about a 1.6$\times$ end-to-end speedup and maintaining VBench and paired-output reconstruction fidelity close to dense InfinityStar.
\end{abstract}
\section{Introduction}
\label{sec:intro}

Diffusion models achieve strong image and video generation quality but require
repeated denoising-network evaluations~\citep{peebles2023dit,
kong2024hunyuanvideo,wan2025}. Visual Autoregressive (VAR) modeling instead
uses next-scale prediction to refine visual tokens from coarse to fine
resolution~\citep{tian2024var}. InfinityStar~\citep{infinitystar} extends this
paradigm to video through Spacetime Autoregressive (STAR) modeling, generating
an image pyramid followed by a sequence of clip pyramids.

\begin{figure}[t]
  \centering
  \includegraphics[width=\columnwidth]{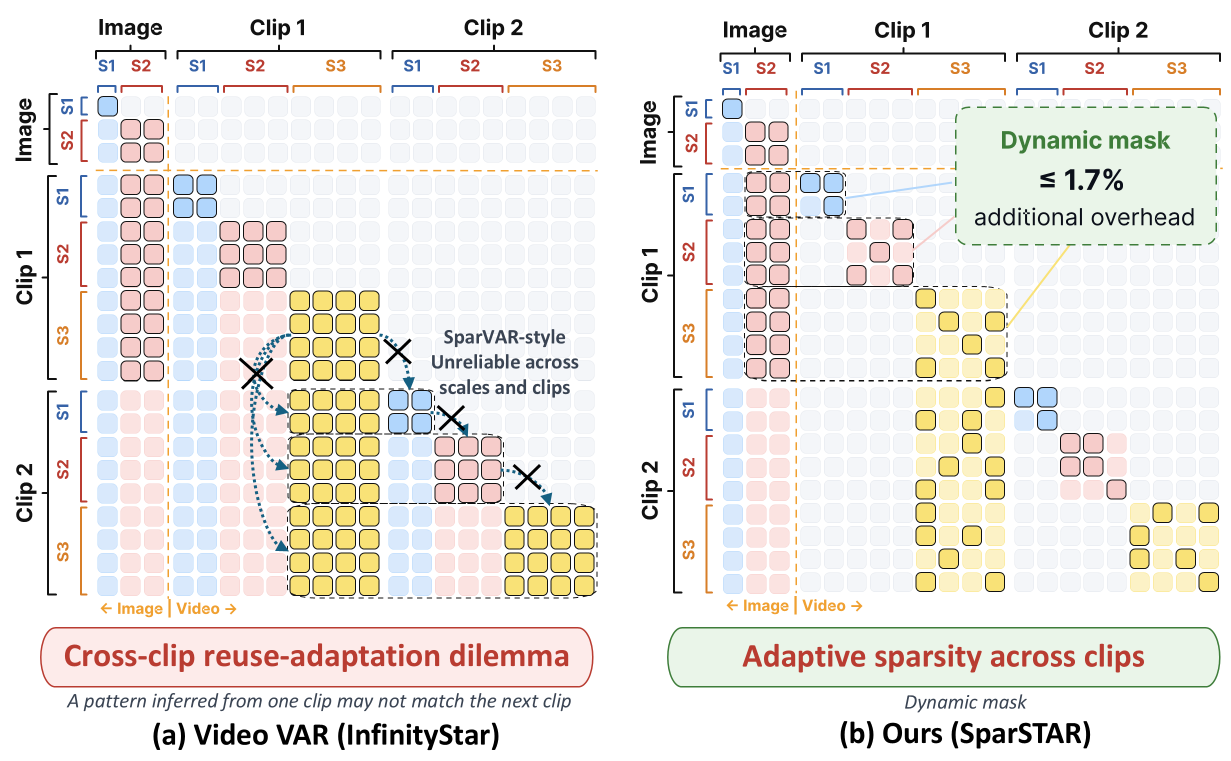}
  \caption{Reused sparse support can miss newly active blocks across scales and clips (a), SparSTAR instead builds a fresh dynamic mask at each sparsified scale with low overhead (b).}
  \label{fig:attention-structure-comparison}
\end{figure}

Despite this promise, attention remains a major bottleneck in VAR models.
Conventional image VAR models~\citep{tian2024var,infinity} accumulate
key--value (KV) pairs from preceding scales, increasing the amount of KV data
processed at high resolutions. For long-video generation, InfinityStar introduces
\emph{Spacetime Sparse Attention} (SSA) to avoid attending to the complete
visual history. As illustrated in Figure~\ref{fig:attention-structure-comparison}(a),
SSA generates clips sequentially and retains only the final scale of the
immediately preceding clip as visual context. Nevertheless, every scale of the
current clip must still attend densely to this preceding-clip context, text
conditioning, and the current-scale tokens. In our evaluated 720p setting, the
final scale alone contains 72{,}000 tokens, leaving substantial attention cost
even after SSA restricts the historical context. This limitation motivates
sparse attention tailored specifically to STAR-based video generation.

Training-free sparse attention~\citep{chen2026sparse,hu2026dfsattn,samuel2026fast,zhang2025fast, zhang2025spargeattention, xia2025training} has been studied extensively for video
diffusion transformers. Radial Attention~\citep{li2025radial} constructs a
static sparse mask from spatiotemporal energy decay, whereas Sparse
VideoGen~\citep{xi2025svg} dynamically profiles each attention head and selects
spatial or temporal sparse patterns. Sparse VideoGen2~\citep{yang2025svg2}
further identifies critical tokens through semantic-aware permutation and
dynamic budget control. These methods are designed around attention over a
fixed-resolution three-dimensional latent grid and repeated denoising
trajectories. Applying them directly to InfinityStar is difficult for two
reasons. First, spatial and temporal priors
derived from a fixed diffusion grid do not directly match the mixed KV context
of SSA. Second, each autoregressive scale is evaluated only once, while its
token resolution, token count, and KV composition change across scales.
Consequently, a sparse pattern profiled at one scale cannot be assumed to
remain valid at later scales.

In image VAR\citep{infinity, tang2025hart}, SparVAR~\citep{sparvar}
introduces cross-scale sparse attention that exploits attention sinks,
cross-scale activation similarity, and spatial locality. SparVAR computes
dense attention at a sparse decision scale and transfers the resulting sparse
pattern to later high-resolution scales through cross-scale index mapping.
Although effective for image VAR, this assumption becomes unreliable in
STAR-based video generation, where the KV context contains the final scale of
the preceding clip and changes at every clip and scale.

To achieve optimal sparse attention in Video VAR, we formulate three core research objectives. First, we analyze the extent to which cross-scale attention patterns remain consistent within Video VAR. Second, we quantify the persistence of the final scale's attention pattern from a prior clip to the next during multi-clip generation. Third, we evaluate the impact of scale distance on performance degradation when cross-scale attention patterns are exploited.

To address these three core research objectives, we present a detailed analysis in Section~\ref{sec:analysis}. Consequently, we uncover that transferring attention patterns across scales and clips in the Video VAR framework induces a significant loss in terms of attention mass. In addition, through extensive experiments varying the range of cross-scale transfer, we find that the performance degradation caused by cross-scale reuse is considerably more severe than the latency overhead introduced by dynamic approaches.

Based on these analysis, we introduce SparSTAR, a training-free
block-sparse attention framework for STAR-based video generation.
Rather than transferring a mask across scales, SparSTAR dynamically constructs a per-head sparse mask from the current query and key activations at each high-cost scale.
It ranks contiguous key blocks and applies a clip-aware policy that keeps
designated context dense while jointly ranking the remaining preceding-clip
and current-scale blocks.
Our cross-scale pattern reuse analysis separates the gap to the token-wise upper bound
into blockization, aggregated-QK scoring, and reuse-induced pattern mismatch terms. This mismatch is
the largest correctable component across most of the evaluated density range,
while fresh aggregated-QK selection consistently approaches the exact block
oracle more closely than cross-scale transfer. A forward-only sparse execution
path keeps dynamic selection practical for an inference pipeline that is
sensitive to additional runtime work.

\noindent Our main contributions are as follows.
\begin{itemize}
  \item Through systematic analysis, we show that attention patterns in InfinityStar do not transfer reliably across scales or clip boundaries, and that reusing them causes substantial attention-mass loss.

  \item We propose a training-free, clip-aware block selector that constructs fresh per-scale and per-head pattern while preserving every token and refinement scale.

  \item We develop a forward-only FlexAttention execution path and evaluate the resulting fidelity--efficiency trade-off on 720p text-to-video and image-to-video generation.
\end{itemize}
\section{Related Work}

\subsection{Visual Generative Models}

Latent diffusion performs iterative denoising in a compressed space, and
Diffusion Transformers replace the U-Net with a transformer backbone
\citep{rombach2022ldm,peebles2023dit}. This fixed-grid, repeated-update
paradigm underpins video models such as HunyuanVideo and
Wan~\citep{kong2024hunyuanvideo,wan2025}. Visual Autoregressive (VAR) models
instead predict increasingly fine token scales~\citep{tian2024var}.
Infinity extends this process to high-resolution images, while InfinityStar
represents video as an image pyramid followed by a sequence of clip pyramids
\citep{infinity,infinitystar}. The changing token resolution and cross-clip
context in InfinityStar create a different acceleration problem from repeated
denoising on a fixed latent grid.

\subsection{Acceleration on Fixed Denoising Grids}

Video-diffusion acceleration~\citep{zhang2026sla2,zhang2026faster, zhang2025sla} exploits repetition across denoising steps and
structure within a fixed spatiotemporal grid. Pyramid Attention Broadcast
reuses attention outputs when consecutive denoising steps change little
\citep{zhao2025pab}. Radial Attention encodes a static decay prior, whereas
Sparse VideoGen profiles heads online to choose spatial or temporal patterns
\citep{li2025radial,xi2025svg}. Sparse VideoGen2 further combines
semantic-aware permutation with a dynamic top-$p$ budget
\citep{yang2025svg2}. These methods differ in whether their patterns are
static, profiled, or reused, but all benefit from a stable grid or repeated
updates. InfinityStar evaluates each autoregressive scale once, while its
query length and visible KV composition change across scales and clips; a
pattern valid at one scale therefore need not transfer to the next.

\subsection{Acceleration across Autoregressive Scales}

Image-VAR acceleration~\citep{li2025stagevar,chen2025collaborative} reduces work by pruning processed tokens, skipping
scale steps, or compressing stored state. FastVAR forwards only pivotal tokens
at large scales, SkipVAR skips selected steps and replaces an unconditional
branch, ScaleKV applies scale-aware KV compression, and SparseVAR excludes
low-frequency tokens while retaining sampled anchors
\citep{guo2025fastvar,li2025skipvar,li2025scalekv,chen2025sparsevar}.
These approaches change the token set, refinement schedule, or KV cache rather
than the attention connections evaluated for every token.

SparVAR directly sparsifies image-VAR attention by transferring a block-selection pattern from a
sparse decision scale to later scales through cross-scale alignment
\citep{sparvar}. This transfer relies on access to historical-scale tokens and
on similarity between block-selection patterns across image scales. Spacetime Sparse Attention masks
those historical scales and exposes only the preceding clip's final scale,
making cross-scale pattern reuse less reliable in STAR-based video generation.

\subsection{Acceleration for STAR-Based Video Generation}

For efficient video VAR~\citep{ye2025fast,li2026packcache, lin2024animatediff}, FastSTAR identifies non-converged tokens from spatial and temporal similarity
and restricts transformer updates to those tokens~\citep{faststar}. SparSTAR
targets a complementary axis: it preserves every token and refinement scale
but reduces the key blocks visited by each query. This distinction motivates
evaluating fidelity to the dense output in addition to raw speed, because
token pruning and attention-edge sparsification perturb the generation process
in different ways.
\begin{figure}[t]
\centering
\includegraphics[width=\columnwidth]{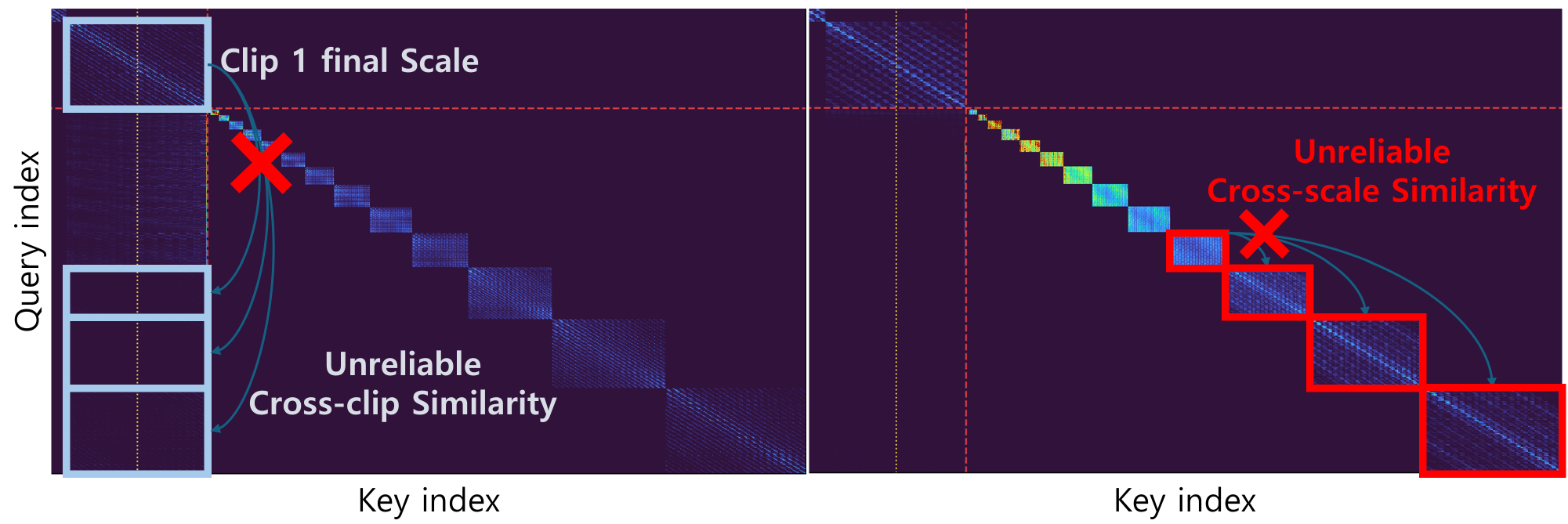}
\par
\makebox[\columnwidth]{%
  \makebox[0.48\columnwidth]{(a)}\hfill
  \makebox[0.48\columnwidth]{(b)}}
\begingroup
\renewcommand{\thefigure}{2}
\caption{Dense-attention patterns across clips and scales in a representative
480p two-clip T2V trace.}
\label{fig:attention-map-structure}
\endgroup
\end{figure}
\section{Preliminaries}
\label{sec:preliminaries}

\subsection{Spacetime Sparse Attention}
\label{sec:prelim-ssa}

InfinityStar~\citep{infinitystar} generates a video through Spacetime Pyramid
Modeling and next-scale prediction, beginning with an image pyramid and then
proceeding through a sequence of clip pyramids. Retaining every previously
generated visual token would cause the historical context to accumulate across
scales and clips. InfinityStar therefore employs \emph{Spacetime Sparse
Attention} (SSA), which masks the earlier visual history and retains only its
final-scale representation. Specifically, Clip~1 conditions on the final
image-pyramid scale, whereas each later clip conditions on the final scale of
the immediately preceding clip. Text conditioning and current-scale tokens
remain visible in both cases. SSA thus differs from conventional image VAR,
which attends to tokens from all preceding scales.

This SSA context layout defines the dense attention region that SparSTAR aims
to sparsify. For clip $c$ at scale $s$, let $\mathbf{Q}^{(c,s)}$ have shape
$L_q^{(c,s)}\times D$ and let $\mathbf{K}^{(c,s)},\mathbf{V}^{(c,s)}$ have
shape $L_k^{(c,s)}\times D$. We consider these head-wise states after the
backbone positional transformations and immediately before the scaled
attention dot product. Let $\mathbf{K}_{\mathrm{ref}}^{(c)}$ denote the
final-scale visual context retained by SSA for clip $c$. The visible key
sequence is
\begin{equation}
  \begin{array}{l}
  \mathbf{K}^{(c,s)}
  =
  \mathrm{Concat}
  \left(
    \mathbf{K}_{\mathrm{ref}}^{(c)},
    \mathbf{K}_{\mathrm{text}},
    \mathbf{K}_{\mathrm{cur}}^{(c,s)}
  \right),
  \\[2pt]
  L_k^{(c,s)}
  =
  L_{\mathrm{ref}}^{(c)}
  +
  L_{\mathrm{text}}
  +
  L_q^{(c,s)}.
  \end{array}
  \label{eq:ssa-layout}
\end{equation}

\begin{figure}[t]
\centering
\includegraphics[width=\columnwidth]{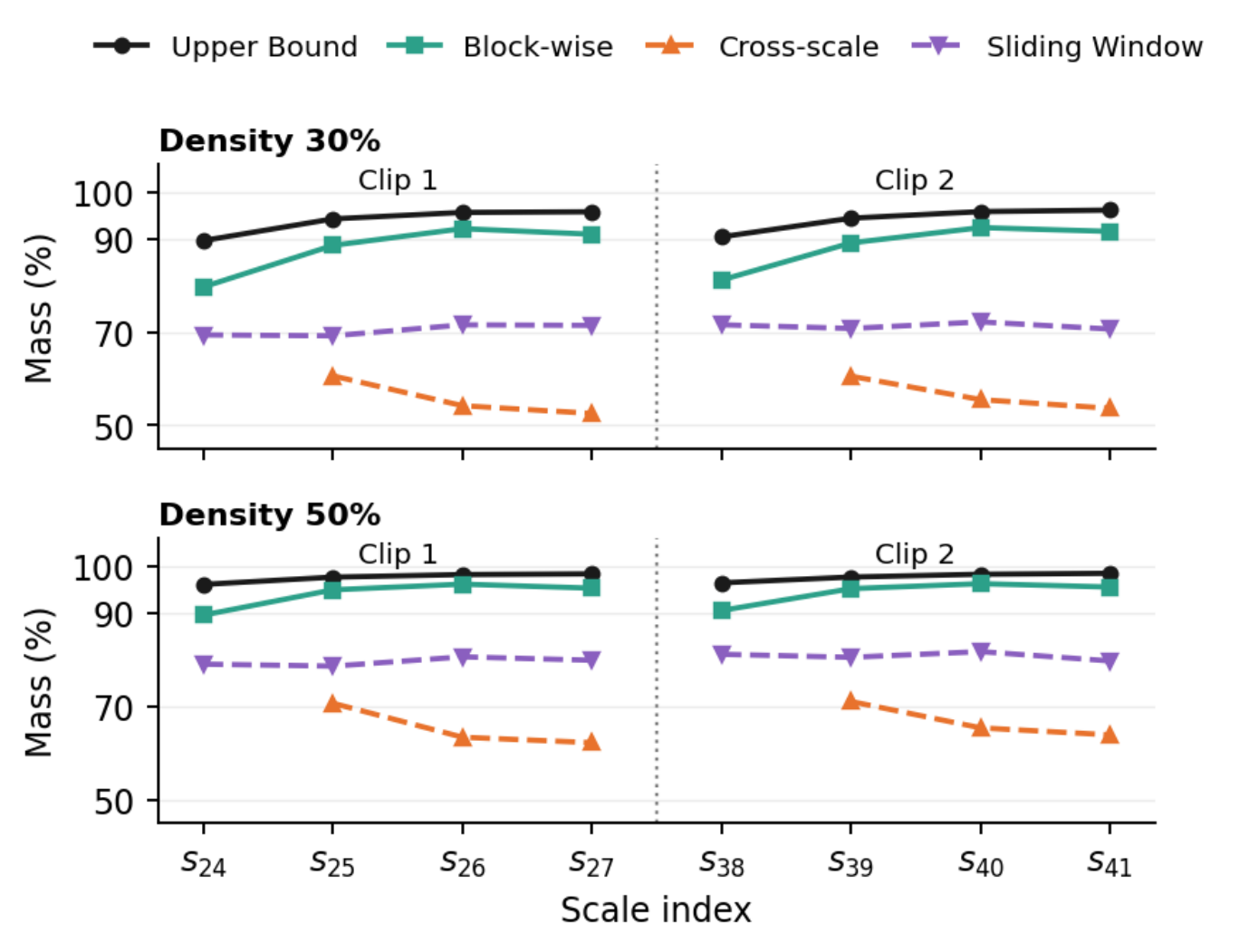}
\begingroup
\renewcommand{\thefigure}{3}
\caption{Scale-wise retained attention mass across Clip~1 and Clip~2 (480p
T2V, 32 prompts, 36 layers, and 32 heads).}
\label{fig:attention-mass-gap}
\endgroup
\end{figure}

The final term is $L_q^{(c,s)}$ because the current-scale visual tokens serve
as keys as well as queries; the value sequence follows the same layout.
Although SSA prevents visual
history from accumulating across all earlier scales and clips, every query
still attends densely to the visible keys. SparSTAR therefore targets the
remaining $L_q^{(c,s)}\times L_k^{(c,s)}$ dense attention region.
\section{Analysis}
\makeatletter
\def\@currentlabel{Analysis}
\makeatother
\label{sec:analysis}

We first inspect attention maps collected while generating a 480p, 10-second T2V sample. Figure~\ref{fig:attention-map-structure} focuses on the transition from Clip~1 to Clip~2, where the final scale of Clip~1 is retained as reference context for Clip~2. Figure~\ref{fig:attention-map-structure}(a) shows that the attention pattern formed over this reference context is not consistent when it is used by Clip~2. Figure~\ref{fig:attention-map-structure}(b) further shows that attention patterns also change across scales within the same clip.

Motivated by these observations, we conduct three analyses to determine how sparse attention can be applied effectively to Video VAR. First, we analyze the extent to which cross-scale attention patterns remain consistent. Second, we quantify the persistence of the final scale's attention pattern from a prior clip to the next during multi-clip generation. Third, we evaluate the impact of scale distance on performance degradation when cross-scale attention patterns are exploited.

\subsection{Cross-Scale Pattern Consistency within a Clip}

Figure~\ref{fig:attention-mass-gap} compares attention-pattern selection at 30\% and 50\% density in the 480p, 10-second T2V setting. The comparison includes token-wise Top-$K$ as an upper bound, Block-wise selection that recomputes Top-$K$ at every scale, cross-scale pattern reuse, and a fixed Sliding Window. For cross-scale reuse, $s_{24}$ in Clip 1 and $s_{38}$ in Clip 2 are used as decision scales, and the attention patterns selected at these scales are reused at the subsequent scales of each clip.

Experimental results show that utilizing attention patterns across scales causes a severe attention mass drop of approximately 35.6\% on average, preserving even less attention mass than a naive Sliding Window approach. In contrast, recomputing Block-wise Top-$K$ at every scale consistently retains attention mass closest to the token-wise upper bound, with an average gap of only 4.64\%. This result demonstrates that recomputing the pattern at each scale is far more effective than reusing a pattern selected at an earlier scale or applying a fixed local pattern.

\subsection{Pattern Persistence across Clip Boundaries}

We next test whether the block-selection pattern from the final scale of
Clip~1 remains valid after its visual tokens become reference context for
Clip~2.

\begin{figure}[t]
\centering
\includegraphics[width=0.95\columnwidth]{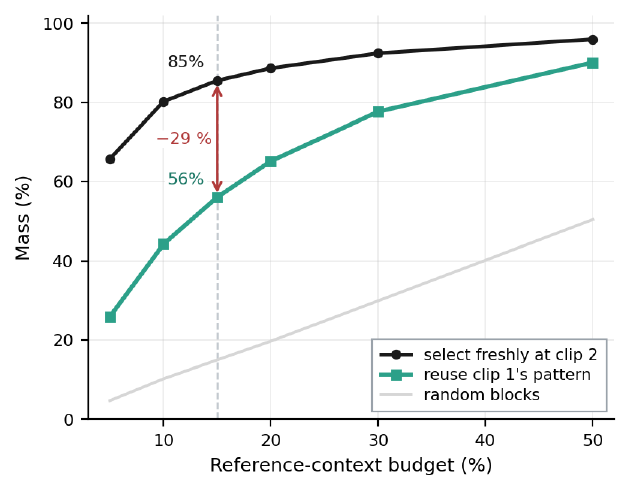}
\caption{Retained attention mass over the preceding-clip reference context.
Fresh selection in Clip~2 consistently retains more mass than reusing the
pattern selected in Clip~1.}
\label{fig:cross-clip-persistence}
\end{figure}

For the same reference context, we compare reusing the block pattern selected while generating Clip~1 with recomputing Block-wise Top-$K$ from the queries of Clip~2.

The reused pattern retains substantially more attention mass than random block selection, showing that some attention structure persists across the clip boundary. However, fresh selection in Clip~2 consistently retains more mass, and the gap becomes larger as the budget decreases. At a 15\% reference-context budget, fresh selection retains 85\% of the attention mass, whereas reuse retains only 56\%, a loss of 29 percentage points. Thus, the pattern selected in Clip~1 remains only partially valid in Clip~2. To use a small reference-context budget effectively, the block selection should be recomputed for each clip rather than reused across clips.

\begin{figure}[!t]
\centering
\includegraphics[width=\columnwidth]{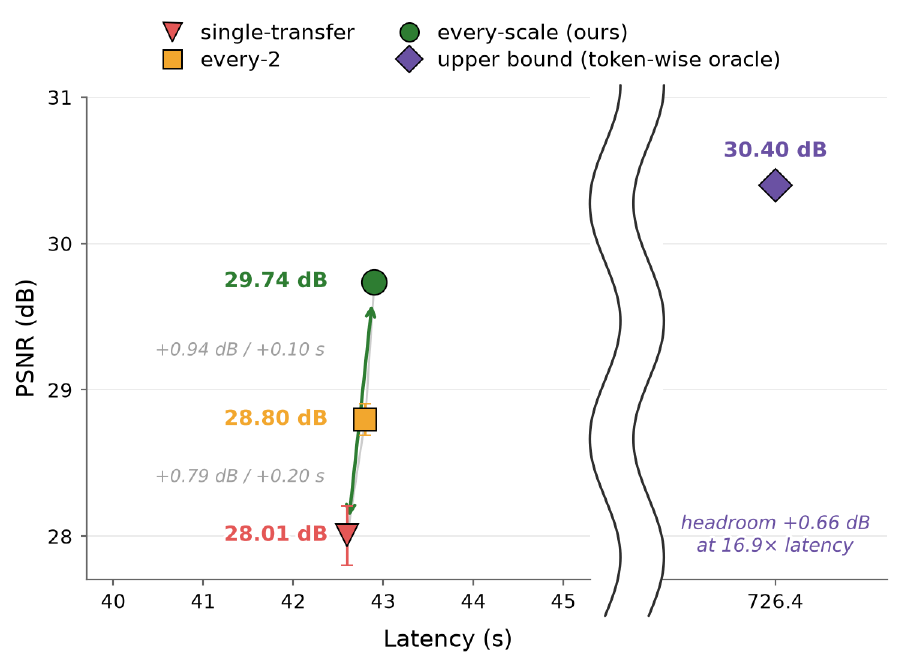}
\caption{PSNR--latency trade-off for recalibration frequency on 720p T2V.
Error bars denote 95\% confidence intervals for PSNR across five seeds.}
\label{fig:recalibration-pareto}
\end{figure}

\subsection{Effect of Scale Distance on Pattern Reuse}

Finally, we analyze how performance changes as an attention pattern selected
at an earlier scale is reused over increasingly distant target scales. We
compare single-transfer, every-2, and every-scale pattern updates on 720p T2V
using 96 prompts and five seeds. All variants share target densities
$(0.625,0.1875,0.1875,0.09375)$ at the four sparsified refinement scales
(native indices $s_{26}$--$s_{29}$). Single-transfer selects a pattern at the
first scale and reuses it thereafter. Every-2 updates the pattern at the first
and third scales, whereas every-scale updates it at all four scales.

Figure~\ref{fig:recalibration-pareto} shows that every-scale pattern updates
improve PSNR by 1.73~dB over single-transfer for only 0.30~s of additional
latency. Every-2 recovers 0.79~dB for 0.20~s.
Moreover, the every-scale result lies close to the upper bound, indicating
that recomputing the pattern at each scale recovers most of the quality
lost through cross-scale reuse. These results indicate that the
performance loss from reusing an earlier-scale pattern increases as the pattern
is carried across more scales, motivating pattern recomputation at every
scale.

\FloatBarrier
\section{Method}
\label{sec:method}

Based on the InfinitySTAR, SparSTAR sparsifies expensive late video scales
without updating model weights or skipping refinement scales. 
SparSTAR first partitions the
query and remaining KV sequence in the order used by the attention kernel. It
then ranks candidate blocks independently for every head from the current
query and key activations. The selected indices are executed through
a forward-only FlexAttention~\citep{dong2024flexattention} path that we
implement and call \textsc{FastFlex}.

\subsection{Aggregated Query--Key Block Selector}
\label{sec:block-selector}

\paragraph{Original-order block partition.}
For a fixed clip, scale, layer, and head, queries and keys are partitioned into
contiguous blocks of $b=128$ tokens with $n_q=\lceil L_q/b\rceil$ and
$n_k=\lceil L_k/b\rceil$. We suppress these context indices on scores,
supports, and masks below, restoring clip or scale indices only where the
selection policy depends on them. The partition follows the attention-kernel
order without aligning blocks to segment boundaries, thereby preserving the
spatiotemporal raster order and matching the tiled execution granularity. A
final partial block is zero-padded and averaged over the nominal block length
$b$.

Let $\mathcal{B}_i^q$ and $\mathcal{B}_j^k$ denote the token-index sets of query block $i$ and key block $j$. Let $\mathcal{I}_{\mathrm{text}}$ and $\mathcal{I}_{\mathrm{ref}}^{(c)}$ denote the text range and the final-scale visual range left unmasked by SSA. Their block sets are $\mathcal{J}_{\mathrm{text}} =\{j\mid\mathcal{B}_j^k\cap\mathcal{I}_{\mathrm{text}}\neq\emptyset\}$ and $\mathcal{J}_{\mathrm{ref}}^{(c)} =\{j\mid\mathcal{B}_j^k\cap \mathcal{I}_{\mathrm{ref}}^{(c)}\neq\emptyset\}$. The intersection definitions also cover segments that cross or share a block boundary.

\paragraph{Per-head aggregated query--key score.}
Let $\widetilde{\mathbf{q}}_{i,r}$ and $\widetilde{\mathbf{k}}_{j,r}$, $r\in\{1,\ldots,b\}$, denote the query and key vectors in a block after zero padding. We compute
$\bar{\mathbf{q}}_i=b^{-1}\sum_r\widetilde{\mathbf{q}}_{i,r}$ and
$\bar{\mathbf{k}}_j=b^{-1}\sum_r\widetilde{\mathbf{k}}_{j,r}$. With
$\tau=D^{-1/2}$, their compatibility is
\begin{equation}
  S_{ij}
  =
  \tau\bar{\mathbf{q}}_i^{\top}\bar{\mathbf{k}}_j.
  \label{eq:block-score}
\end{equation}
By bilinearity, this dot product equals the average pairwise pre-softmax
compatibility between tokens in two full blocks. For partial blocks, zero
padding scales the aggregated vector magnitude with block occupancy. Since softmax
is nonlinear, $S_{ij}$ serves only as a ranking score rather than an estimate
of post-softmax attention mass.

Scores are computed independently for every head from the current $\mathbf{Q}$ and $\mathbf{K}$ without sharing patterns across heads or scales. The parameter-free block score reduces token-level query--key work by a factor of approximately $b^2=16{,}384$, excluding aggregation and selection overhead. Across the sparsified late scales, selector arithmetic is 0.016\% of the corresponding dense query--key arithmetic.

\begin{table*}[t]
\centering
\begingroup
\small
\setlength{\tabcolsep}{4.5pt}
\renewcommand{\arraystretch}{1.10}

\begin{tabularx}{\textwidth}{
  >{\raggedright\arraybackslash}p{0.14\textwidth}
  >{\raggedright\arraybackslash}p{0.16\textwidth}
  *{5}{>{\centering\arraybackslash}X}}
\toprule
\textbf{Category}
& \textbf{Method}
& \textbf{PSNR} ($\uparrow$)
& \textbf{SSIM} ($\uparrow$)
& \textbf{LPIPS} ($\downarrow$)
& \textbf{VBench} ($\uparrow$)
& \textbf{Speedup} ($\uparrow$) \\
\midrule
\multicolumn{7}{c}{\textbf{720p Text-to-Video (T2V)}} \\
\addlinespace[2pt]
Baseline & InfinityStar & -- & -- & -- & 83.79 & 1.00$\times$ \\
\midrule
\addlinespace[1pt]
\multirow{4}{*}{Token Reduction} & SparseVAR & 26.59 & 0.789 & 0.228 & 82.13 & 1.44$\times$ \\
& FastVAR  & 25.00 & 0.757 & 0.266 & 81.19 & 1.61$\times$ \\
& ToMe     & 26.47 & 0.790 & 0.227 & 81.60 & 1.76$\times$ \\
& FastSTAR & 28.30 & 0.828 & 0.184 & 82.96 & 1.92$\times$ \\
\midrule
\rowcolor{blue!10}
Sparse Attention & \textbf{SparSTAR} & \textbf{31.22} & \textbf{0.898} & \textbf{0.065} & \textbf{83.83} & 1.60$\times$ \\
\midrule
\multicolumn{7}{c}{\textbf{720p Image-to-Video (I2V)}} \\
\addlinespace[2pt]
Baseline & InfinityStar & -- & -- & -- & 81.04 & 1.00$\times$ \\
\midrule
\addlinespace[1pt]
\multirow{4}{*}{Token Reduction} & SparseVAR & 23.86 & 0.754 & 0.249 & 78.30 & 1.56$\times$ \\
& FastVAR  & 22.20 & 0.711 & 0.292 & 74.83 & 1.82$\times$ \\
& ToMe     & 23.93 & 0.757 & 0.248 & 78.25 & 1.57$\times$ \\
& FastSTAR & 25.65 & 0.809 & 0.195 & 80.01 & 2.01$\times$ \\
\midrule
\rowcolor{blue!10}
Sparse Attention & \textbf{SparSTAR} & \textbf{27.96} & \textbf{0.870} & \textbf{0.080} & \textbf{80.96} & 1.62$\times$ \\
\bottomrule
\end{tabularx}
\endgroup
\caption{Quality and efficiency for 5-second, 81-frame 720p T2V and I2V
generation. All T2V evaluations use the \texttt{refined\_prompt} strings
distributed with InfinityStar. Dashes indicate non-applicable metrics.}
\label{tab:t2v-i2v-main-results}
\end{table*}

\subsection{Clip-Dependent Retained and Selectable Blocks}
\label{sec:clip-conditioned-selection}

SparSTAR defines retained and selectable block sets separately for Clip~1 and later clips. For Clip~1, text and final image-pyramid blocks remain dense, and only current-scale blocks are selected. For Clip~$c\geq2$, only text blocks remain dense, while preceding-clip final-scale and current-scale blocks are jointly ranked under the same budget.

Formally, the always-retained set is
\begin{equation}
  \mathcal{J}_{\mathrm{keep}}^{(c)}
  =
  \left\{
  \begin{array}{ll}
    \mathcal{J}_{\mathrm{text}}
    \cup
    \mathcal{J}_{\mathrm{ref}}^{(1)},
    & c=1,\\
    \mathcal{J}_{\mathrm{text}},
    & c\geq2,
  \end{array}
  \right.
  \label{eq:clip-keep-set}
\end{equation}
and the selectable set is
\begin{equation}
  \mathcal{J}_{\mathrm{sel}}^{(c)}
  =
  \{1,\ldots,n_k\}
  \setminus
  \mathcal{J}_{\mathrm{keep}}^{(c)},
  \qquad
  n_{\mathrm{sel}}^{(c)}
  =
  |\mathcal{J}_{\mathrm{sel}}^{(c)}|.
  \label{eq:clip-selectable-set}
\end{equation}
Only blocks in $\mathcal{J}_{\mathrm{sel}}^{(c)}$ require ranking, while $\mathcal{J}_{\mathrm{keep}}^{(c)}$ remains dense without a score.

\paragraph{Scale-wise density schedule.}
We use separate density lookup functions for Clip~1 and later clips, indexed by query length as $\rho_{c,s}=\rho_c(L_q^{(c,s)})$. Scales with the same token count share a fixed density within each clip group. The non-uniform schedule assigns higher density to cheaper late scales and the most aggressive budget to the 72{,}000-token final scale. This final scale accounts for 74.6\% of attention cost among the sparsified scales and exhibits the most concentrated pattern. The selected block count is
\begin{equation}
  k_{c,s}
  =
  \min\!\left(
    n_{\mathrm{sel}}^{(c)},
    \max\!\left(
      1,
      \left\lceil
        \rho_{c,s}n_{\mathrm{sel}}^{(c)}
      \right\rceil
    \right)
  \right).
  \label{eq:scale-budget}
\end{equation}
The reported 720p setting sparsifies the four final refinement scales,
ordered by increasing token count. In the backbone's native indexing these
are $s_{26}$ through $s_{29}$, with target densities
$(\rho_{26},\rho_{27},\rho_{28},\rho_{29})=
(0.625,0.1875,0.1875,0.09375)$.
For a fixed layer and head at clip $c$ and scale $s$, the retained key-block set of
query block $i$ is
\begin{equation}
  \mathcal{A}_i
  =
  \mathcal{J}_{\mathrm{keep}}^{(c)}
  \cup
  \mathrm{TopKIdx}_{j\in\mathcal{J}_{\mathrm{sel}}^{(c)}}
  \left(S_{ij},k_{c,s}\right).
  \label{eq:retained-support}
\end{equation}
Later clips impose no minimum quota on preceding-clip final-scale blocks, so a
head may select none when current-scale blocks receive higher scores. The fixed
schedule is applied to every prompt.

\begin{figure}[t]
\centering
\includegraphics[width=\columnwidth]{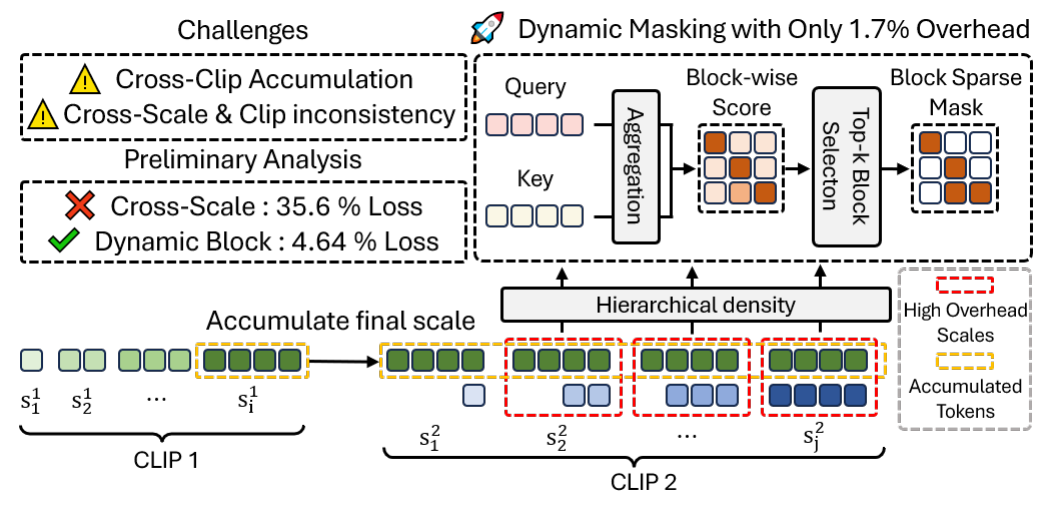}
\caption{Overview of SparSTAR, a training-free block-sparse attention framework for video autoregressive generation that reduces attention latency while maintaining video quality.}
\label{fig:sparstar-method-overview}
\end{figure}

\subsection{Forward-Only Sparse Execution}
\label{sec:sparse-execution}

Let $\pi_q(t)$ and $\pi_k(u)$ map query and key tokens to their block indices. The selected token-level mask is
\begin{equation}
  M_{tu}
  =
  \mathbf{1}\!\left[
    \pi_k(u)
    \in
    \mathcal{A}_{\pi_q(t)}
  \right].
  \label{eq:final-sparse-mask}
\end{equation}
SSA has already masked tokens outside the allowed context, so SparSTAR only removes additional attention connections. The executed mask combines the SparSTAR mask with the backbone padding and boundary masks.

FlexAttention represents the \textsc{BlockMask} and
skips masked tiles. Its mask stores, for each query block, the number and
indices of visible key blocks; backward execution additionally requires the
reverse mapping from key blocks to query blocks. Because generation is
inference-only, \textsc{FastFlex} writes only the forward fields
\texttt{kv\_num\_blocks} and \texttt{kv\_indices} from
$\{\mathcal{A}_i\}_{i=1}^{n_q}$ and omits the backward-only
\texttt{q\_num\_blocks} and \texttt{q\_indices}. 
This avoids constructing reverse indexing structures that inference never consumes, with the measured selection and mask-construction overhead reported in Section~\ref{sec:experiments}.

\subsection{Complexity and Scope}
\label{sec:complexity}

Dense attention costs $O(L_qL_kD)$ per layer and head.
SparSTAR executes attention only over the always-retained and selected KV
blocks. Its attention cost therefore scales approximately as
$O(\rho_{\mathrm{eff}}L_qL_kD)$, where $\rho_{\mathrm{eff}}$ is the fraction
of KV blocks retained. Selection operates on aggregated 128-token blocks, so its
cost grows with the number of query--key block pairs rather than token pairs.

SparSTAR reads fewer KV blocks without evicting or compressing the KV cache. Image-pyramid generation, early video scales, and non-attention modules remain unchanged. SparSTAR therefore preserves all tokens and refinement scales while targeting late-scale attention.
\section{Experiments}
\label{sec:experiments}

\begin{table*}[t]
\centering
\begingroup
\small
\setlength{\tabcolsep}{1.5pt}
\renewcommand{\arraystretch}{1.08}

\begin{tabularx}{\textwidth}{
  >{\raggedright\arraybackslash}p{0.20\textwidth}
  *{5}{>{\centering\arraybackslash}X}
  !{\vrule width 0.4pt}
  *{5}{>{\centering\arraybackslash}X}}
\toprule
\multirow{2}{*}{\textbf{Method}}
& \multicolumn{5}{c}{\textbf{720p 5s Text-to-Video (T2V)}}
& \multicolumn{5}{c}{\textbf{720p 5s Image-to-Video (I2V)}} \\
\cmidrule(lr){2-6}\cmidrule(lr){7-11}
& \makecell{\textbf{PSNR}\\($\uparrow$)}
& \makecell{\textbf{SSIM}\\($\uparrow$)}
& \makecell{\textbf{LPIPS}\\($\downarrow$)}
& \makecell{\textbf{VBench}\\($\uparrow$)}
& \makecell{\textbf{Speedup}\\($\uparrow$)}
& \makecell{\textbf{PSNR}\\($\uparrow$)}
& \makecell{\textbf{SSIM}\\($\uparrow$)}
& \makecell{\textbf{LPIPS}\\($\downarrow$)}
& \makecell{\textbf{VBench}\\($\uparrow$)}
& \makecell{\textbf{Speedup}\\($\uparrow$)} \\
\midrule
SparSTAR
& 31.22 & 0.898 & 0.065 & 83.83 & 1.60$\times$
& 27.96 & 0.870 & 0.080 & 80.96 & 1.62$\times$ \\
SparSTAR + Scale-skip
& 30.13 & 0.881 & 0.086 & 83.15 & 2.27$\times$
& 27.28 & 0.845 & 0.104 & 78.78 & 2.20$\times$ \\
SparSTAR + Guidance-off
& 29.97 & 0.883 & 0.077 & 83.55 & 2.50$\times$
& 27.32 & 0.859 & 0.090 & 79.56 & 2.50$\times$ \\
\bottomrule
\end{tabularx}
\endgroup
\caption{Compatibility of SparSTAR with complementary acceleration on 720p
T2V and I2V. All T2V evaluations use the \texttt{refined\_prompt} strings
distributed with InfinityStar. Guidance-off skips the unconditional CFG branch
at all four sparsified refinement scales, while Scale-skip omits the final scale.}
\label{tab:complementary-acceleration}
\end{table*}

\subsection{Experimental Setup}

\paragraph{Models and evaluation tasks.}
We evaluate SparSTAR on the 8B InfinityStar~\citep{infinitystar} for
text-to-video (T2V) and image-to-video (I2V) generation. We reports
720p results, while 480p results and fine-grained VBench scores are provided in the supplementary material. The
720p evaluation uses 5-second, 81-frame videos following the evaluation
protocol of InfinityStar adopted by FastSTAR~\citep{faststar}. We use 
InfinityStar as the reference for reconstruction metrics and normalize its
end-to-end latency to $1.00\times$. We measure the InfinityStar and
SparSTAR VBench scores directly using matched prompts and seeds.

\paragraph{Baselines and source of results.}
We compare SparSTAR against SparseVAR~\citep{chen2025sparsevar},
FastVAR~\citep{guo2025fastvar}, ToMe~\citep{bolya2023tome}, and
FastSTAR~\citep{faststar}. The accelerated-baseline results in
Table~\ref{tab:t2v-i2v-main-results} are quoted directly from the FastSTAR
paper rather than obtained from our own re-evaluation; their speedups are
therefore not hardware-matched to ours and are treated as contextual published
results. FastSTAR extends the image-oriented token-pruning and token-merging
methods across the spatiotemporal pyramid while retaining the default
InfinityStar generation hyperparameters.
Because cross-scale transfer by SparVAR requires historical tokens masked by
Spacetime Sparse Attention, we assess reused block-selection pattern separately in the
cross-scale pattern reuse analysis above.

\paragraph{Quality and efficiency metrics.}
Following FastSTAR~\citep{faststar}, PSNR, SSIM, and LPIPS quantify
reconstruction fidelity and perceptual similarity against paired dense
InfinityStar outputs. These reconstruction metrics measure agreement with the
dense reference rather than absolute perceptual quality. VBench~\citep{huang2024vbench} evaluates 5-second,
81-frame videos over all 16 dimensions. The headline results in
Tables~\ref{tab:t2v-i2v-main-results}, \ref{tab:complementary-acceleration},
and~\ref{tab:480p-long-video-results} use the complete official VBench suite:
each dimension is evaluated on its designated prompt subset with the five
default seeds per prompt. Dimension-balanced subsets, including the 96-prompt
set with six prompts per dimension, are reserved for the analyses and
ablations whose prompt scopes are reported separately in the supplementary
material.
Dashes denote metrics that are not applicable to the dense reference.

\paragraph{Implementation and hardware.}
Experiments use BF16 on two NVIDIA H100 NVL GPUs with 94~GB of memory each,
PyTorch 2.5.1+cu124, and Triton 3.1.0. SparSTAR uses 128-token blocks and
sparsifies only high-resolution refinement scales. SparSTAR and dense
InfinityStar use matched tasks, resolutions, prompts, and seeds.

\begin{table}[t]
\centering
\begingroup
\scriptsize
\setlength{\tabcolsep}{1.2pt}
\renewcommand{\arraystretch}{1.04}
\begin{tabularx}{\columnwidth}{
  >{\raggedright\arraybackslash}p{0.21\columnwidth}
  *{5}{>{\centering\arraybackslash}X}}
\toprule
\multicolumn{6}{c}{\textbf{480p, 10s Text-to-Video (T2V)}} \\
\cmidrule(lr){1-6}
\textbf{Method}
& \makecell{\textbf{PSNR}\\($\uparrow$)}
& \makecell{\textbf{SSIM}\\($\uparrow$)}
& \makecell{\textbf{LPIPS}\\($\downarrow$)}
& \makecell{\textbf{VBench}\\($\uparrow$)}
& \makecell{\textbf{Speedup}\\($\uparrow$)} \\
\midrule
InfinityStar
& -- & -- & -- & 83.64 & 1.00$\times$ \\
SparSTAR
& 24.39 & 0.7759 & 0.145 & 82.77 & 1.31$\times$ \\
\midrule
\multicolumn{6}{c}{\textbf{480p, 10s Image-to-Video (I2V)}} \\
\cmidrule(lr){1-6}

InfinityStar
& -- & -- & -- & 77.83 & 1.00$\times$ \\
SparSTAR
& 27.00 & 0.8425 & 0.096 & 77.37 & 1.36$\times$ \\
\bottomrule
\end{tabularx}
\endgroup
\caption{Quality and efficiency for 480p, 10-second T2V and I2V generation.}
\vspace{-8pt}
\label{tab:480p-long-video-results}
\end{table}

\subsection{Main Results}

\paragraph{Quality preservation.}
Table~\ref{tab:t2v-i2v-main-results} shows that SparSTAR preserves the
generation behavior of dense InfinityStar while improving paired-output
reconstruction fidelity over the published accelerated baselines. Unlike
token-reduction methods, SparSTAR does not discard latent tokens or omit their
transformer updates: every token proceeds through every refinement scale, but
each query reads fewer key blocks. This preserves the dense refinement
trajectory more directly. Relative to the reported FastSTAR results, SparSTAR
improves PSNR by 2.92~dB on T2V and 2.31~dB on I2V, while its VBench scores
remain within 0.04 and 0.08 points of the dense reference. SSIM and LPIPS show
the same paired-output fidelity trend.

\paragraph{End-to-end efficiency.}
In Table~\ref{tab:t2v-i2v-main-results},
SparSTAR provides 1.60$\times$ T2V and 1.62$\times$ I2V speedups while
preserving all tokens and refinement scales. Within our matched final-scale
benchmark, the sparse operator is 7.08$\times$ faster than
FlashAttention-2~\citep{dao2023flashattention2}, while selection and mask
construction add less than 1.7\% end-to-end latency. Unchanged early scales
and non-attention modules account for the smaller pipeline-level gain.

\paragraph{Compatibility with complementary acceleration.}
Guidance-off means skipping the unconditional branch of classifier-free
guidance (CFG)~\citep{ho2022cfg} at all four sparsified refinement scales,
while Scale-skip means omitting the final refinement scale.
Table~\ref{tab:complementary-acceleration} shows that Guidance-off provides
the stronger trade-off, reaching 2.50$\times$ speedup on both tasks while
preserving higher VBench and reconstruction fidelity than Scale-skip.
Guidance-off removes an unconditional branch while leaving the conditional
refinement hierarchy intact; Scale-skip instead removes the final
high-resolution update itself. Branch-level reduction therefore complements
SparSTAR with less disruption to the generated representation.

\subsection{Long-Video Evaluation}
\begin{figure}[t]
\centering

\includegraphics[width=\columnwidth]{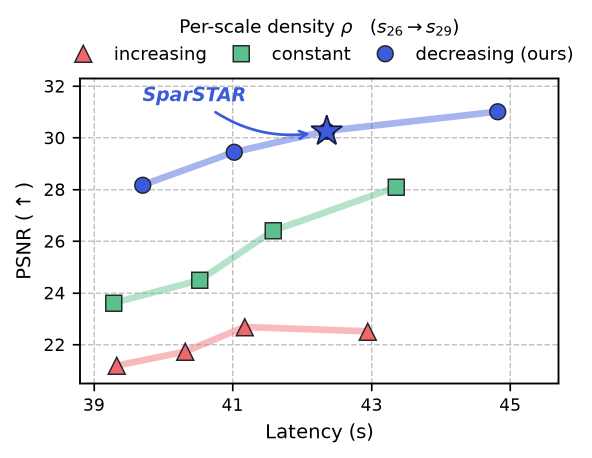}
\caption{Comparison of increasing, constant, and decreasing density schedules
across the sparsified refinement scales.}

\label{fig:density-ablation}
\end{figure}

In multi-clip generation, every scale of Clip~2 attends to the final-scale
reference retained from Clip~1. This increases the key sequence and attention
cost. We generate Clip~1 with dense attention. We then compare dense attention
and SparSTAR only on Clip~2. Table~\ref{tab:480p-long-video-results} reports
1.31$\times$ and 1.36$\times$ speedups for T2V and I2V. The corresponding
VBench reductions are only 0.87 and 0.46 points. These results show that
SparSTAR reduces reference-context overhead while largely preserving
multi-clip generation quality.

\subsection{Ablation Study}
\subsubsection{Density Ratios}\mbox{}\par

To study how the per-scale density ratio affects computation time and
generation quality, we compare three schedules in
Figure~\ref{fig:density-ablation}. The decreasing schedule used by
SparSTAR lowers density as the scale grows. The increasing schedule reverses
this order. The constant schedule uses one density throughout. At similar
latency, the decreasing schedule achieves the highest PSNR. Early scales
establish global structure. Removing their connections causes errors that
later refinement cannot recover. Late scales contain more tokens, so stronger
sparsity at these scales yields larger savings. The increasing schedule
reverses this priority. The constant schedule ignores it. These results
support SparSTAR's decreasing density schedule.

\FloatBarrier
\section{Conclusion}
We presented SparSTAR, a training-free dynamic block-sparse attention method
that further sparsifies the context visible under InfinityStar's Spacetime
Sparse Attention. Trace analysis shows that recomputing aggregated-QK block selection at
each sparsified scale avoids substantial attention-mass loss from reused
cross-scale masks.
Across 720p T2V and I2V evaluation, SparSTAR improves reconstruction fidelity
over the published accelerated baselines while providing substantial operator
and end-to-end speedups. The method leaves early scales and non-attention
modules unchanged and does not compress the KV cache. Future work should
evaluate longer multi-clip generation, broader model families, sensitivity to
block size and density schedules, the contribution of the clip-aware policy,
and the interaction between sparse pattern and temporal consistency.

\clearpage
\bibliography{reference}

@inproceedings{tian2024var,
  title        = {Visual Autoregressive Modeling: Scalable Image Generation via Next-Scale Prediction},
  author       = {Keyu Tian and Yi Jiang and Zehuan Yuan and Bingyue Peng and Liwei Wang},
  booktitle    = {Advances in Neural Information Processing Systems (NeurIPS)},
  year         = {2024}
}

@inproceedings{infinity,
  title        = {Infinity: Scaling Bitwise AutoRegressive Modeling for High-Resolution Image Synthesis},
  author       = {Jian Han and Jinlai Liu and Yi Jiang and Bin Yan and Yuqi Zhang and Zehuan Yuan and Bingyue Peng and Xiaobing Liu},
  booktitle    = {Proceedings of the IEEE/CVF Conference on Computer Vision and Pattern Recognition (CVPR)},
  year         = {2025}
}

@inproceedings{infinitystar,
  title        = {InfinityStar: Unified Spacetime AutoRegressive Modeling for Visual Generation},
  author       = {Jinlai Liu and Jian Han and Bin Yan and Hui Wu and Fengda Zhu and Xing Wang and Yi Jiang and Bingyue Peng and Zehuan Yuan},
  booktitle    = {Advances in Neural Information Processing Systems (NeurIPS)},
  year         = {2025}
}

@inproceedings{peebles2023dit,
  title        = {Scalable Diffusion Models with Transformers},
  author       = {William Peebles and Saining Xie},
  booktitle    = {Proceedings of the IEEE/CVF International Conference on Computer Vision (ICCV)},
  year         = {2023}
}

@inproceedings{rombach2022ldm,
  title        = {High-Resolution Image Synthesis with Latent Diffusion Models},
  author       = {Robin Rombach and Andreas Blattmann and Dominik Lorenz and Patrick Esser and Bj{\"o}rn Ommer},
  booktitle    = {Proceedings of the IEEE/CVF Conference on Computer Vision and Pattern Recognition (CVPR)},
  year         = {2022}
}

@article{kong2024hunyuanvideo,
  title        = {HunyuanVideo: A Systematic Framework For Large Video Generative Models},
  author       = {Weijie Kong and Qi Tian and Zijian Zhang and Rox Min and Zuozhuo Dai and Jin Zhou and Jiangfeng Xiong and Xin Li and Bo Wu and Jianwei Zhang and Kathrina Wu and Qin Lin and Junkun Yuan and Yanxin Long and Aladdin Wang and Andong Wang and others},
  journal      = {arXiv preprint arXiv:2412.03603},
  year         = {2024}
}

@article{wan2025,
  title        = {Wan: Open and Advanced Large-Scale Video Generative Models},
  author       = {Team Wan and Ang Wang and Baole Ai and Bin Wen and Chaojie Mao and Chen-Wei Xie and Di Chen and Feiwu Yu and Haiming Zhao and Jianxiao Yang and Jingren Zhou and others},
  journal      = {arXiv preprint arXiv:2503.20314},
  year         = {2025}
}

@inproceedings{zhao2025pab,
  title        = {Real-Time Video Generation with Pyramid Attention Broadcast},
  author       = {Xuanlei Zhao and Xiaolong Jin and Kai Wang and Yang You},
  booktitle    = {International Conference on Learning Representations (ICLR)},
  year         = {2025}
}

@inproceedings{xi2025svg,
  title        = {Sparse VideoGen: Accelerating Video Diffusion Transformers with Spatial-Temporal Sparsity},
  author       = {Haocheng Xi and Shuo Yang and Yilong Zhao and Chenfeng Xu and Muyang Li and Xiuyu Li and Yujun Lin and Han Cai and Jintao Zhang and Dacheng Li and Jianfei Chen and Ion Stoica and Kurt Keutzer and Song Han},
  booktitle    = {International Conference on Machine Learning (ICML)},
  year         = {2025}
}

@inproceedings{yang2025svg2,
  title        = {Sparse VideoGen2: Accelerate Video Generation with Sparse Attention via Semantic-Aware Permutation},
  author       = {Shuo Yang and Haocheng Xi and Yilong Zhao and Muyang Li and Jintao Zhang and Han Cai and Yujun Lin and Xiuyu Li and Chenfeng Xu and Jianfei Chen and Song Han and Kurt Keutzer and Ion Stoica},
  booktitle    = {Advances in Neural Information Processing Systems (NeurIPS)},
  year         = {2025}
}

@inproceedings{li2025radial,
  title        = {Radial Attention: $O(n\log n)$ Sparse Attention with Energy Decay for Long Video Generation},
  author       = {Xingyang Li and Muyang Li and Tianle Cai and Haocheng Xi and Shuo Yang and Yujun Lin and Lvmin Zhang and Songlin Yang and Jinbo Hu and Kelly Peng and Maneesh Agrawala and Ion Stoica and Kurt Keutzer and Song Han},
  booktitle    = {Advances in Neural Information Processing Systems (NeurIPS)},
  year         = {2025}
}

@article{faststar,
  title        = {FastSTAR: Spatiotemporal Token Pruning for Efficient Autoregressive Video Synthesis},
  author       = {Sungwoong Yune and Suheon Jeong and Joo-Young Kim},
  journal      = {arXiv preprint arXiv:2603.07192},
  year         = {2026}
}

@inproceedings{dong2024flexattention,
  title        = {Flex Attention: A Programming Model for Generating Optimized Attention Kernels},
  author       = {Juechu Dong and Boyuan Feng and Driss Guessous and Yanbo Liang and Horace He},
  booktitle    = {Proceedings of Machine Learning and Systems (MLSys)},
  year         = {2025}
}

@inproceedings{dao2023flashattention2,
  title        = {FlashAttention-2: Faster Attention with Better Parallelism and Work Partitioning},
  author       = {Tri Dao},
  booktitle    = {International Conference on Learning Representations (ICLR)},
  year         = {2024}
}

@article{ho2022cfg,
  title        = {Classifier-Free Diffusion Guidance},
  author       = {Jonathan Ho and Tim Salimans},
  journal      = {arXiv preprint arXiv:2207.12598},
  year         = {2022}
}

@inproceedings{huang2024vbench,
  title        = {{VBench}: Comprehensive Benchmark Suite for Video Generative Models},
  author       = {Ziqi Huang and Yinan He and Jiashuo Yu and Fan Zhang and Chenyang Si and Yuming Jiang and Yuanhan Zhang and Tianxing Wu and Qingyang Jin and Nattapol Chanpaisit and Yaohui Wang and Xinyuan Chen and Limin Wang and Dahua Lin and Yu Qiao and Ziwei Liu},
  booktitle    = {Proceedings of the IEEE/CVF Conference on Computer Vision and Pattern Recognition (CVPR)},
  year         = {2024}
}

@inproceedings{bolya2023tome,
  title        = {Token Merging: Your {ViT} but Faster},
  author       = {Daniel Bolya and Cheng-Yang Fu and Xiaoliang Dai and Peizhao Zhang and Christoph Feichtenhofer and Judy Hoffman},
  booktitle    = {International Conference on Learning Representations (ICLR)},
  year         = {2023}
}

@inproceedings{guo2025fastvar,
  title        = {FastVAR: Linear Visual Autoregressive Modeling via Cached Token Pruning},
  author       = {Hang Guo and Yawei Li and Taolin Zhang and Jiangshan Wang and Tao Dai and Shu-Tao Xia and Luca Benini},
  booktitle    = {Proceedings of the IEEE/CVF International Conference on Computer Vision (ICCV)},
  year         = {2025}
}

@inproceedings{sparvar,
  title        = {SparVAR: Exploring Sparsity in Visual AutoRegressive Modeling for Training-Free Acceleration},
  author       = {Zekun Li and Ning Wang and Tongxin Bai and Changwang Mei and Peisong Wang and Shuang Qiu and Jian Cheng},
  booktitle    = {Proceedings of the IEEE/CVF Conference on Computer Vision and Pattern Recognition (CVPR)},
  year         = {2026}
}

@inproceedings{li2025scalekv,
  title        = {Memory-Efficient Visual Autoregressive Modeling with Scale-Aware KV Cache Compression},
  author       = {Kunjun Li and Zigeng Chen and Cheng-Yen Yang and Jenq-Neng Hwang},
  booktitle    = {Advances in Neural Information Processing Systems (NeurIPS)},
  year         = {2025}
}

@article{li2025skipvar,
  title        = {SkipVAR: Accelerating Visual Autoregressive Modeling via Adaptive Frequency-Aware Skipping},
  author       = {Jiajun Li and Yue Ma and Xinyu Zhang and Qingyan Wei and Songhua Liu and Linfeng Zhang},
  journal      = {arXiv preprint arXiv:2506.08908},
  year         = {2025}
}

@inproceedings{chen2025sparsevar,
  title        = {Frequency-Aware Autoregressive Modeling for Efficient High-Resolution Image Synthesis},
  author       = {Zhuokun Chen and Jugang Fan and Zhuowei Yu and Bohan Zhuang and Mingkui Tan},
  booktitle    = {Proceedings of the IEEE/CVF International Conference on Computer Vision (ICCV)},
  year         = {2025}
}

@inproceedings{tang2025hart,
  title={Hart: Efficient visual generation with hybrid autoregressive transformer},
  author={Tang, Haotian and Wu, Yecheng and Yang, Shang and Xie, Enze and Chen, Junsong and Chen, Junyu and Zhang, Zhuoyang and Cai, Han and Lu, Yao and Han, Song},
  booktitle={International Conference on Learning Representations},
  volume={2025},
  pages={68413--68432},
  year={2025}
}

@article{li2025stagevar,
  title={StageVAR: Stage-Aware Acceleration for Visual Autoregressive Models},
  author={Li, Senmao and Wang, Kai and Khan, Salman and Khan, Fahad Shahbaz and Yang, Jian and Wang, Yaxing},
  journal={arXiv preprint arXiv:2512.16483},
  year={2025}
}

@inproceedings{chen2025collaborative,
  title={Collaborative decoding makes visual auto-regressive modeling efficient},
  author={Chen, Zigeng and Ma, Xinyin and Fang, Gongfan and Wang, Xinchao},
  booktitle={Proceedings of the Computer Vision and Pattern Recognition Conference},
  pages={23334--23344},
  year={2025}
}

@article{ye2025fast,
  title={Fast autoregressive video generation with diagonal decoding},
  author={Ye, Yang and Guo, Junliang and Wu, Haoyu and He, Tianyu and Pearce, Tim and Rashid, Tabish and Hofmann, Katja and Bian, Jiang},
  journal={arXiv preprint arXiv:2503.14070},
  year={2025}
}

@article{li2026packcache,
  title={PackCache: A Training-Free Acceleration Method for Unified Autoregressive Video Generation via Compact KV-Cache},
  author={Li, Kunyang and Shah, Mubarak and Shang, Yuzhang},
  journal={arXiv preprint arXiv:2601.04359},
  year={2026}
}

@article{lin2024animatediff,
  title={Animatediff-lightning: Cross-model diffusion distillation},
  author={Lin, Shanchuan and Yang, Xiao},
  journal={arXiv preprint arXiv:2403.12706},
  year={2024}
}

@inproceedings{hu2026dfsattn,
  title={DFSAttn: Dynamic Fine-grained Sparse Attention for Efficient Video Generation},
  author={Hu, Jie and Gao, Zixiang and He, Yutong and Yuan, Kun},
  booktitle={Forty-third International Conference on Machine Learning},
  year={2026}
}

@article{samuel2026fast,
  title={Fast autoregressive video diffusion and world models with temporal cache compression and sparse attention},
  author={Samuel, Dvir and Tzachor, Issar and Levy, Matan and Green, Michael and Chechik, Gal and Ben-Ari, Rami},
  journal={arXiv preprint arXiv:2602.01801},
  year={2026}
}

@article{zhang2025fast,
  title={Fast video generation with sliding tile attention},
  author={Zhang, Peiyuan and Chen, Yongqi and Su, Runlong and Ding, Hangliang and Stoica, Ion and Liu, Zhengzhong and Zhang, Hao},
  journal={arXiv preprint arXiv:2502.04507},
  year={2025}
}

@article{zhang2025spargeattention,
  title={Spargeattention: Accurate and training-free sparse attention accelerating any model inference},
  author={Zhang, Jintao and Xiang, Chendong and Huang, Haofeng and Wei, Jia and Xi, Haocheng and Zhu, Jun and Chen, Jianfei},
  journal={arXiv preprint arXiv:2502.18137},
  year={2025}
}

@inproceedings{xia2025training,
  title={Training-free and adaptive sparse attention for efficient long video generation},
  author={Xia, Yifei and Ling, Suhan and Fu, Fangcheng and Wang, Yujie and Li, Huixia and Xiao, Xuefeng and Cui, Bin},
  booktitle={Proceedings of the IEEE/CVF International Conference on Computer Vision},
  pages={15982--15993},
  year={2025}
}

@article{zhang2026sla2,
  title={Sla2: Sparse-linear attention with learnable routing and qat},
  author={Zhang, Jintao and Wang, Haoxu and Jiang, Kai and Zheng, Kaiwen and Jiang, Youhe and Stoica, Ion and Chen, Jianfei and Zhu, Jun and Gonzalez, Joseph E},
  journal={arXiv preprint arXiv:2602.12675},
  year={2026}
}

@article{zhang2026faster,
  title={Faster video diffusion with trainable sparse attention},
  author={Zhang, Peiyuan and Chen, Yongqi and Huang, Haofeng and Lin, Will and Liu, Zhengzhong and Stoica, Ion and Xing, Eric and Zhang, Hao},
  journal={Advances in Neural Information Processing Systems},
  volume={38},
  pages={152509--152534},
  year={2026}
}

@article{zhang2025sla,
  title={Sla: Beyond sparsity in diffusion transformers via fine-tunable sparse-linear attention},
  author={Zhang, Jintao and Wang, Haoxu and Jiang, Kai and Yang, Shuo and Zheng, Kaiwen and Xi, Haocheng and Wang, Ziteng and Zhu, Hongzhou and Zhao, Min and Stoica, Ion and others},
  journal={arXiv preprint arXiv:2509.24006},
  year={2025}
}

@inproceedings{chen2026sparse,
  title={Sparse-vdit: Unleashing the power of sparse attention to accelerate video diffusion transformers},
  author={Chen, Pengtao and Zeng, Xianfang and Zhao, Maosen and Shen, Mingzhu and Cheng, Wei and Yu, Gang and Chen, Tao},
  booktitle={Proceedings of the AAAI Conference on Artificial Intelligence},
  volume={40},
  pages={2957--2965},
  year={2026}
}

\clearpage
\setcounter{page}{1}
\setcounter{figure}{7}
\setcounter{table}{3}
\setcounter{equation}{7}
\setcounter{secnumdepth}{2}

\setcounter{topnumber}{3}
\setcounter{bottomnumber}{2}
\setcounter{totalnumber}{4}
\setcounter{dbltopnumber}{3}
\renewcommand{\topfraction}{0.92}
\renewcommand{\bottomfraction}{0.80}
\renewcommand{\dbltopfraction}{0.92}
\renewcommand{\textfraction}{0.06}
\renewcommand{\floatpagefraction}{0.70}
\renewcommand{\dblfloatpagefraction}{0.70}
\setlength{\dblfloatsep}{3pt plus 1pt minus 1pt}
\setlength{\dbltextfloatsep}{6pt plus 1pt minus 2pt}

\newcommand{\MainTabMain}{1}
\newcommand{\MainTabLongVideo}{3}
\newcommand{\MainFigMassGap}{3}

\let\SuppOriginalLabel\label
\renewcommand{\label}[1]{%
  \pdfdest name{#1} xyz%
  \SuppOriginalLabel{#1}%
}
\newcommand{\SuppContentsLink}[2]{%
  \leavevmode
  \pdfstartlink attr{/Border[0 0 0]} goto name{#1}%
  #2%
  \pdfendlink
}
\newcommand{\SuppSecRow}[2]{%
  \SuppContentsLink{#1}{\textbf{\ref{#1}}}
  & \SuppContentsLink{#1}{\textbf{#2}}
  & \SuppContentsLink{#1}{\textbf{\pageref{#1}}} \\%
}
\newcommand{\SuppSubRow}[2]{%
  & \SuppContentsLink{#1}{\ref{#1}\quad #2}\dotfill
  & \SuppContentsLink{#1}{\pageref{#1}} \\%
}

\appendix
\onecolumn
\section*{Supplementary Material}

\vspace{1.5em}
\noindent{\large\scshape Contents}\par
\vspace{1.5em}

\begingroup
\centering
\renewcommand{\arraystretch}{1.25}
\begin{tabularx}{0.98\textwidth}{@{}p{0.06\textwidth}Xr@{}}
\SuppSecRow{sec:supp-dataset-evaluation-details}{Dataset and Evaluation Details}
\SuppSubRow{sec:supp-vbench-configuration}{VBench Configuration and Aggregation}
\SuppSubRow{sec:supp-prompt-seed-protocol}{Prompt Set, Seeds, and InfinityStar Refined T2V Prompts}
\SuppSubRow{sec:supp-i2v-conditioning}{I2V Conditioning Images}
\SuppSubRow{sec:supp-reconstruction-metrics}{Reconstruction Metrics and Dense Pairing}
\SuppSubRow{sec:supp-analysis-configuration}{Analysis Prompt Set and Backbone Structure}
\SuppSubRow{sec:supp-video-scale-configurations}{Video, Clip, Frame, and Scale Configurations}
\SuppSubRow{sec:supp-sparsity-schedule}{Sparsity Schedule and Later-Clip Policy}
\SuppSubRow{sec:supp-baseline-protocols}{Published Baseline Protocols}
\SuppSubRow{sec:supp-evaluation-scope}{Evaluation Scope: Full Suite vs.\ Analysis Subset}
\addlinespace[6pt]
\SuppSecRow{sec:supp-implementation-release}{Implementation and Release Details}
\SuppSubRow{sec:supp-hardware-software}{Hardware and Software}
\SuppSubRow{sec:supp-code-availability}{Code Availability}
\addlinespace[6pt]
\SuppSecRow{sec:supp-additional-quantitative-results}{Additional Quantitative Results}
\SuppSubRow{sec:supp-480p-quality-efficiency}{480p Five-Second Quality and Efficiency}
\SuppSubRow{sec:supp-fine-grained-vbench}{Fine-Grained VBench Results}
\SuppSubRow{sec:supp-vbench-ci}{Per-Dimension Differences with Confidence Intervals}
\addlinespace[6pt]
\SuppSecRow{sec:supp-additional-ablation}{Additional Ablation Study}
\SuppSubRow{sec:supp-selection-strategy-analysis}{Selection Strategy Analysis}
\SuppSubRow{sec:supp-block-size-ablation}{Block-Size Ablation}
\addlinespace[6pt]
\SuppSecRow{sec:supp-additional-qualitative-results}{Additional Qualitative Results}
\SuppSubRow{sec:supp-qualitative-comparisons}{Qualitative Comparisons}
\end{tabularx}
\endgroup

\clearpage
\twocolumn
\raggedbottom

\section{Dataset and Evaluation Details}
\label{sec:supp-dataset-evaluation-details}

\subsection{VBench Configuration and Aggregation}
\label{sec:supp-vbench-configuration}
We use the original 16-dimension VBench protocol rather than VBench++ or
VBench-2.0. The evaluated T2V dimensions are subject consistency, background
consistency, temporal flickering, motion smoothness, dynamic degree, aesthetic
quality, imaging quality, object class, multiple objects, human action, color,
spatial relationship, scene, appearance style, temporal style, and overall
consistency. The seven quality dimensions form the Quality Score, and the nine
semantic dimensions form the Semantic Score. Following the official aggregate
script, the normalized total score is computed as
\begin{equation}
  \mathrm{VBench}_{\mathrm{total}}
  = 100\,\frac{4\,\mathrm{Quality}+\mathrm{Semantic}}{5}.
  \label{eq:supp-vbench-total}
\end{equation}
The I2V evaluation reports nine dimensions: I2V subject, I2V background,
subject consistency, background consistency, temporal flickering, motion
smoothness, aesthetic quality, imaging quality, and dynamic degree.

\paragraph{Version record.}
All VBench numbers use the \texttt{vbench} package version 0.1.5 (the original
16-dimension benchmark, not VBench++ or VBench-2.0), with the prompt and
metadata file \texttt{VBench\_full\_info.json} shipped inside that package
(946 prompts).
Per-dimension prompt-pool sizes in this file are 72 for subject consistency,
motion smoothness, and dynamic degree; 75 for temporal flickering; 79 for
object class; 82 for multiple objects; 84 for spatial relationship; 85 for
color; 86 for background consistency and scene; 90 for appearance style; 93
for aesthetic quality, imaging quality, and overall consistency; and 100 for
human action and temporal style. Evaluation is invoked through
\texttt{vbench.VBench.evaluate} with
\texttt{mode=\textquotesingle vbench\_standard\textquotesingle}, one dimension
per call, so each dimension reads exactly its own prompt subset. Aggregate
scores use the normalization ranges and dimension weights from the VBench
repository's \texttt{scripts/constant.py} (dynamic degree weighted $0.5$, all
other dimensions $1.0$), combined as in
Equation~(\ref{eq:supp-vbench-total}). The I2V evaluation uses the
\texttt{vbench2\_beta\_i2v} module from the same package.

\subsection{Prompt Set, Seeds, and InfinityStar Refined T2V Prompts}
\label{sec:supp-prompt-seed-protocol}
For the dimension-balanced T2V analyses, we use six prompts per dimension,
giving 96 prompt assignments across the 16 dimensions. This subset is distinct
from the complete official VBench suite used for the headline results, as
detailed in Section~\ref{sec:supp-evaluation-scope}. The analysis prompt set is
fixed before comparing dense InfinityStar and SparSTAR.

\paragraph{Seeds and subset construction.}
Seeds are deterministic and require no seed manifest: sample $j$ of a prompt
uses seed $41+j$, so the five default seeds are $41,\ldots,45$
(\texttt{SEED\_BASE}${}=41$, seed ${}={}$
\texttt{SEED\_BASE}${}+{}$sample index). The dimension-balanced subsets are
built by a recorded procedure rather than by hand. Each dimension takes six
prompts chosen \emph{evenly spaced} over that dimension's pool. Adjacent
VBench indices are often near-duplicates (for
example, \textit{``a bicycle''} and \textit{``a bicycle and a car''}), so
taking the first $n$ would collapse the diversity of the subset. The resulting
96 dimension-specific prompt assignments are evaluated independently, which
at five seeds gives 480 videos per method.

\paragraph{InfinityStar refined T2V prompts.}
We do not run a prompt rewriter. All T2V evaluations consume
\texttt{evaluation/VBench\_rewrited\_prompt.json} exactly as distributed with
the public InfinityStar release, so there is no rewriter model, system
instruction, user template, decoding configuration, or post-processing step
of ours to disclose. Reproducing our setup requires only that file, which is
already public. It holds 946 entries, one per VBench prompt, each carrying
\texttt{prompt\_en}, its \texttt{dimension} list, and a
\texttt{refined\_prompt}. The generator is fed the
\texttt{refined\_prompt} string, while filenames, manifests, and all VBench
bookkeeping keep the official \texttt{prompt\_en}, which makes the outputs
discoverable by the standard evaluator. For example, \texttt{prompt\_en}
\textit{``In a still frame, a stop sign''} maps to the
\texttt{refined\_prompt} \textit{``In a still frame, a stop sign stands
prominently against a clear blue sky backdrop. The octagonal red sign with
bold white letters is positioned slightly off-center, drawing attention with
its vivid color.\ldots''} All T2V methods use this same refined-prompt set.

\subsection{I2V Conditioning Images}
\label{sec:supp-i2v-conditioning}
Conditioning images come from the official VBench-I2V image suite, downloaded
as the benchmark's own \texttt{crop.zip} release (2.46\,GB) and used
unmodified; we redistribute no images. The suite ships pre-cropped variants at
three aspect ratios (1:1, 16:9, and 3:2) with 355 images each, and we use the
\textbf{16:9} set for all I2V experiments, matching the generated video's
aspect ratio. Because the crops are supplied by the benchmark, no resize rule,
crop coordinates, or interpolation kernel of ours enters the pipeline. Each
image is identified by its \texttt{file\_name} in the accompanying
\texttt{i2v-bench-info.json}, which also records the source \texttt{url},
content \texttt{type}, original resolution, the two crop rectangles, and the
\texttt{caption} used as the text condition. The release will list those 355
identifiers together with the suite revision rather than the image files
themselves. Pixels are read as 8-bit RGB in $[0,255]$ with no additional
normalization beyond the backbone's own preprocessing.

\begin{table*}[!t]
\centering
\begingroup
\small
\setlength{\tabcolsep}{4pt}
\renewcommand{\arraystretch}{1.15}
\begin{tabularx}{\textwidth}{@{}lcc>{\raggedright\arraybackslash}X>{\raggedright\arraybackslash}X>{\raggedright\arraybackslash}X@{}}
\toprule
\textbf{Setting}
& \textbf{Duration}
& \textbf{Frames}
& \textbf{Clip Composition}
& \textbf{Native Scale Information}
& \textbf{SparSTAR Configuration} \\
\midrule
720p T2V/I2V
& 5~s
& 81
& One appearance frame followed by one 80-frame clip pyramid at 16~fps.
& Four final refinement scales are $s_{26}$--$s_{29}$; the final scale contains 72{,}000 query tokens.
& Retained densities $0.625$, $0.1875$, $0.1875$, and $0.09375$ at $s_{26}$--$s_{29}$; all earlier scales remain dense. \\
480p T2V/I2V
& 5~s
& 81
& One appearance frame followed by one 80-frame clip pyramid at 16~fps.
& The backbone has 28 native scales and 14 video-tower scales. The final four
scales are $s_{24}$--$s_{27}$ with 5{,}040, 11{,}520, 20{,}640, and 31{,}800
query tokens.
& Retained densities $0.625$, $0.1875$, $0.1875$, and $0.09375$ at
$s_{24}$--$s_{27}$; all earlier scales remain dense. \\
480p T2V/I2V
& 10~s
& 161
& One appearance frame followed by two 80-frame clip pyramids; Clip~2 conditions on the final scale of Clip~1.
& Clip~1 ends at $s_{27}$. The final four scales of Clip~2 are
$s_{38}$--$s_{41}$, with 5{,}040, 11{,}520, 20{,}640, and 31{,}800 query
tokens, respectively.
& Clip~1 remains dense, and its final scale is retained as reference context
for Clip~2. SparSTAR is applied only at $s_{38}$--$s_{41}$ with densities
$0.625$, $0.1875$, $0.1875$, and $0.09375$, respectively. \\
\bottomrule
\end{tabularx}
\endgroup
\caption{Video, clip, frame, scale, and SparSTAR configurations used in the
reported experiments.}
\label{tab:supp-video-scale-configuration}
\end{table*}

\subsection{Reconstruction Metrics and Dense Pairing}
\label{sec:supp-reconstruction-metrics}
PSNR is reported in decibels, whereas SSIM and LPIPS are unitless. They measure
agreement with paired dense InfinityStar outputs rather than absolute video
quality. Dense InfinityStar and SparSTAR use matched tasks, resolutions,
prompts, I2V conditioning images, frame counts, and generation seeds. Frames
are compared at matching temporal indices and spatial resolution without
motion compensation or temporal realignment.

\paragraph{Reconstruction-metric definitions.}
All three metrics are computed on decoded RGB frames, not on luminance, in the
range $[0,255]$ with 8-bit quantization, at matched temporal indices and native
spatial resolution. PSNR is computed \emph{per frame} and then averaged over
frames, rather than pooled over all pixels of the clip. By Jensen's inequality,
this reports slightly higher values than the pooled variant (for the 480p
10\,s setting, 24.39\,dB per frame against 24.07\,dB pooled). SSIM uses
\texttt{scikit-image}'s \texttt{structural\_similarity} at its defaults: a
$7\times7$ uniform (not Gaussian) window, per-channel evaluation averaged over
channels, $\mathrm{ddof}=1$ covariance normalization,
\texttt{data\_range}${}=255$, and the 3-pixel border discarded. LPIPS uses the
\texttt{lpips} package with the AlexNet backbone on frames scaled to $[-1,1]$,
averaged over frames. Reconstruction metrics are computed on every clip of the
corresponding run, not on a separately sampled subset; for the 720p full-suite
T2V comparison, this is all 4{,}730 clips per method. As an implementation check, a
GPU reimplementation of these estimators agrees with the reference
implementation to $3.7\times10^{-6}$\,dB for PSNR and to the printed precision
for SSIM and LPIPS. SSIM must be accumulated in double precision:
$\mathrm{E}[X^2]-\mathrm{E}[X]^2$ cancels two quantities of order $255^2$,
and evaluating it in single precision shifts SSIM by up to
$9\times10^{-3}$ per clip.

\subsection{Analysis Prompt Set and Backbone Structure}
\label{sec:supp-analysis-configuration}
The 480p attention analyses use 32 prompts and the 8B InfinityStar backbone
with 36 transformer layers and 32 attention heads. Thus, each evaluated
prompt and scale contains 1{,}152 layer--head attention instances before
aggregation. The analysis records both Clip~1 and Clip~2 behavior; for the
cross-scale reuse experiment, $s_{24}$ in Clip~1 and $s_{38}$ in Clip~2 are
the decision scales.

\paragraph{Analysis aggregation.}
The 32 prompts are drawn by the same dimension-balanced procedure described in
Section~\ref{sec:supp-prompt-seed-protocol}. The analysis covers all 36
transformer layers and all 32 attention heads on the 10-second configuration
(Clip~1 own finest scales $s_{24}$--$s_{27}$ and Clip~2 own finest scales
$s_{38}$--$s_{41}$). Statistics are macro-averaged: first, an unweighted mean
is taken over causal source-to-later-target scale pairs, layers, and heads
within each prompt; then the bootstrap resamples the 32 prompt-level means.
Confidence intervals are 95\% percentile-bootstrap intervals with 20{,}000
resamples.

\subsection{Video, Clip, Frame, and Scale Configurations}
\label{sec:supp-video-scale-configurations}
Table~\ref{tab:supp-video-scale-configuration} summarizes the video, clip,
frame, scale, and sparsity configurations used in the reported experiments.
All three settings start from a single appearance frame and then generate
80-frame clip pyramids at 16~fps, so the only structural difference between them
is how many clips follow that frame: the five-second settings generate one, and
the ten-second setting generates two, with Clip~2 conditioning on the final
scale of Clip~1. In the five-second settings, SparSTAR sparsifies the final four
refinement scales: $s_{26}$--$s_{29}$ at 720p and $s_{24}$--$s_{27}$ at 480p.
In the ten-second setting, Clip~1 remains dense, its final scale is retained as
reference context for Clip~2, and SparSTAR is applied only to Clip~2 scales
$s_{38}$--$s_{41}$. All sparsified scales use retained densities
$(0.625, 0.1875, 0.1875, 0.09375)$ in ascending scale order.

\begin{table*}[!t]
\centering
\begingroup
\small
\setlength{\tabcolsep}{4pt}
\renewcommand{\arraystretch}{1.15}
\begin{tabularx}{\textwidth}{@{}l>{\raggedright\arraybackslash}X>{\raggedright\arraybackslash}X>{\raggedright\arraybackslash}X@{}}
\toprule
\textbf{Method}
& \textbf{Source of Reported Numbers}
& \textbf{Hardware and Protocol}
& \textbf{Comparability} \\
\midrule
Dense InfinityStar / SparSTAR
& Re-evaluated in our environment with matched tasks, resolutions, prompts, conditioning inputs, and seeds.
& Ubuntu 22.04.5, BF16, PyTorch~2.5.1+cu124, and Triton~3.1.0 on H100 NVL GPUs
(94~GB each). Each video occupies one GPU at batch size one. End-to-end timing
includes the complete generation pipeline between CUDA synchronizations, after
one discarded warm-up video.
& Direct paired comparison. Latency is the mean over the timed evaluation
videos. \\
FastSTAR
& Values are transcribed from the FastSTAR paper rather than re-evaluated by us.
& The setting uses a single NVIDIA H100 (80~GB) for 720p, 5-second, 81-frame videos. End-to-end latency includes the text encoder and VAE decoder. The paper uses InfinityStar's distributed \texttt{refined\_prompt} strings for T2V and 10 videos per VBench dimension for reconstruction metrics.
& This is a contextual result. It is neither hardware- nor sample-matched to ours. \\
SparseVAR / FastVAR / ToMe
& Values are transcribed from the FastSTAR implementations of these methods on InfinityStar.
& FastSTAR extends the original image-oriented methods across the spatiotemporal pyramid while retaining default InfinityStar generation hyperparameters and evaluates them on a single H100 (80~GB).
& Contextual published results. They should not be interpreted as reproductions on our hardware or as the original papers' native image-generation protocols. \\
\bottomrule
\end{tabularx}
\endgroup
\caption{Provenance and comparability of the reported baseline results.}
\label{tab:supp-baseline-protocols}
\end{table*}

\subsection{Sparsity Schedule and Later-Clip Policy}
\label{sec:supp-sparsity-schedule}
For the reported 720p setting, SparSTAR sparsifies native scales
$s_{26}$--$s_{29}$ with retained densities
$(0.625, 0.1875, 0.1875, 0.09375)$, respectively. Earlier scales remain dense,
and the 72{,}000-token final scale uses the smallest retained density because
it accounts for 74.6\% of the attention cost among the sparsified scales while
concentrating most of its attention mass on a small subset of keys. The same
fixed schedule is applied to every prompt.

\paragraph{Later-clip schedule.}
For the reported ten-second evaluation, Clip~1 is generated with dense
attention. Its final scale is retained as reference context for every scale of
Clip~2. SparSTAR is applied only to the four finest scales
of Clip~2, $s_{38}$--$s_{41}$, with retained densities
$(0.625, 0.1875, 0.1875, 0.09375)$, respectively. Text blocks remain dense,
while the preceding-clip reference blocks and current-scale blocks are jointly
ranked under the corresponding density budget.

\subsection{Published Baseline Protocols}
\label{sec:supp-baseline-protocols}
Table~\ref{tab:supp-baseline-protocols} distinguishes results measured in our
environment from published values. Our dense InfinityStar and SparSTAR runs
use the same H100 NVL hardware, prompts, conditioning inputs, and seeds, so
their quality and latency form a paired comparison. The FastSTAR, SparseVAR,
FastVAR, and ToMe results are reported by FastSTAR and use a different H100
variant. FastSTAR reports reconstruction metrics on 10 videos per VBench
dimension, but its exact evaluation code and prompt subset were not available
to us; we instead evaluate all official prompts with five seeds for broader
coverage. Published baseline values are therefore contextual rather than
hardware- or sample-matched comparisons with our results.

\subsection{Evaluation Scope: Full Suite vs.\ Analysis Subset}
\label{sec:supp-evaluation-scope}
Two different prompt scopes are used in this paper, and every table states which one it
follows.

\paragraph{Headline results use the full VBench suite.}
For clarity, references to the 96-prompt configuration denote a
dimension-balanced subset used only for analyses and ablations, not the
evaluation scope of the headline results.
Tables~\MainTabMain{}--\MainTabLongVideo{} of the main paper and the
fine-grained per-dimension results of Section~\ref{sec:supp-fine-grained-vbench}
are computed with the official
\texttt{vbench\_standard} protocol over the complete benchmark: each of the 16 dimensions
is scored on its own official prompt subset, with the five default seeds per prompt. At
720p, T2V uses all 946 official prompts with five seeds per prompt, while I2V
uses all 355 official conditioning images with five seeds each. Each dimension
consumes its designated subset. Both the dense reference and SparSTAR are
evaluated under this identical protocol, so the per-dimension means are
directly comparable. The aggregate VBench column follows the official
normalization constants and dimension weights, combining the seven quality
dimensions and nine semantic dimensions as
$(4\,\mathrm{Quality} + \mathrm{Semantic})/5$.

\paragraph{Analyses and ablations use dimension-balanced subsets.}
The attention-level analyses and the ablation studies do not re-run the full suite. They
use dimension-balanced prompt subsets --- primarily 96 prompt assignments (six prompts per
dimension across the 16 dimensions) --- so that many arms can be generated and compared
under identical conditions within the compute budget. Because these tables compare arms
that are all evaluated on the same subset with the same seeds, the comparisons remain
internally valid; only their absolute values are not directly comparable with the
full-suite numbers in Tables~\MainTabMain{}--\MainTabLongVideo{}.
Table~\ref{tab:supp-evaluation-scope} states which scope each table follows.

\begin{table*}[!t]
\centering
\begingroup
\footnotesize
\setlength{\tabcolsep}{5pt}
\renewcommand{\arraystretch}{1.15}
\begin{tabularx}{\textwidth}{@{}>{\raggedright\arraybackslash}X>{\raggedright\arraybackslash}Xcc@{}}
\toprule
\textbf{Results} & \textbf{Evaluation set}
& \textbf{T2V videos/arm} & \textbf{I2V videos/arm} \\
\midrule
Main Tables~\MainTabMain--\MainTabLongVideo;
Tables~\ref{tab:supp-480p-results}--\ref{tab:supp-vbench-480p}
& Full VBench / VBench-I2V suite
& 946\,$\times$\,5 & 355\,$\times$\,5 \\
Table~\ref{tab:supp-vbench-ci}
& Separate 720p T2V batch
& Varies by dimension & -- \\
\midrule
Tables~\ref{tab:supp-selection-strategy},~\ref{tab:supp-block-size-ablation}
& Dimension-balanced T2V subset
& 96\,$\times$\,5 ${}={}$ 480 & -- \\
\bottomrule
\end{tabularx}
\endgroup
\caption{Evaluation scope and number of generated videos per method arm. Our
dense InfinityStar--SparSTAR comparisons use matched evaluation sets and
seeds; published baseline results follow their original evaluation protocols.}
\label{tab:supp-evaluation-scope}
\end{table*}
\section{Implementation and Release Details}
\label{sec:supp-implementation-release}

\subsection{Hardware and Software}
\label{sec:supp-hardware-software}
All measurements run on Ubuntu 22.04.5 LTS (kernel 5.15.0) with two NVIDIA H100
NVL GPUs (94\,GB each, driver 565.57.01), an Intel Xeon Silver 4514Y host (64
logical cores, 125\,GiB RAM), Python 3.11, PyTorch 2.5.1+cu124 (cuDNN 9.1.0),
and Triton 3.1.0, in BF16. SparSTAR uses 128-token blocks and a forward-only
FlexAttention path at the selected high-resolution scales. Each generation
occupies one GPU; batch size is one video, and reported latency is per video.
End-to-end timing brackets the complete generation pipeline---the text
encoder, all refinement scales, and the VAE decoder---between
\texttt{torch.cuda.synchronize()} calls. Sparse runs generate one warm-up video
before timing begins because the first call pays FlexAttention's
\texttt{torch.compile} cost; the warm-up video is discarded and never enters
any statistic. The reported statistic is the mean over timed videos.

\subsection{Code Availability}
\label{sec:supp-code-availability}
All code required to reproduce generation, sparse selection, FlexAttention
execution, preprocessing, metric computation, and table aggregation will be
made publicly available upon publication. The release will include prompt and
seed manifests, I2V image identifiers, environment files, model configuration,
per-scale density schedules, timing scripts, and raw per-video evaluation
outputs.
\section{Additional Quantitative Results}
\label{sec:supp-additional-quantitative-results}

\subsection{480p Five-Second Quality and Efficiency}
\label{sec:supp-480p-quality-efficiency}
Table~\ref{tab:supp-480p-results} summarizes reconstruction quality, overall
VBench performance, and end-to-end speedup for 480p, five-second T2V and I2V
generation. PSNR, SSIM, and LPIPS measure agreement with paired dense
InfinityStar outputs, whereas VBench evaluates the perceptual and semantic
quality of the generated videos. Speedup is measured relative to dense
InfinityStar inference. The picture is the same shape as at 720p. SparSTAR
stays closer to the dense output than FastSTAR by 2.60~dB on T2V and 2.28~dB on
I2V, and roughly halves LPIPS on both tasks, while its VBench score lands within
0.41 points of dense InfinityStar on T2V and 1.12 points on I2V. It buys that
fidelity at a smaller speedup than FastSTAR, 1.30$\times$ against 1.47$\times$,
since 480p offers less attention work to remove than 720p does.

\begin{table}[!tb]
\begingroup
\centering
\small
\setlength{\tabcolsep}{2pt}
\renewcommand{\arraystretch}{1.10}

\begin{tabularx}{\linewidth}{
  @{}>{\raggedright\arraybackslash}p{0.22\linewidth}
  *{5}{>{\centering\arraybackslash}X}@{}}
\toprule
\textbf{Method}
& \makecell{\textbf{PSNR}\\($\uparrow$)}
& \makecell{\textbf{SSIM}\\($\uparrow$)}
& \makecell{\textbf{LPIPS}\\($\downarrow$)}
& \makecell{\textbf{VBench}\\($\uparrow$)}
& \makecell{\textbf{Speedup}\\($\uparrow$)} \\
\midrule
\multicolumn{6}{c}{\textbf{480p Text-to-Video (T2V)}} \\
\addlinespace[2pt]
InfinityStar & -- & -- & -- & 83.89 & 1.00$\times$ \\
\midrule
\addlinespace[1pt]
FastSTAR & 22.96 & 0.697 & 0.262 & -- & 1.47$\times$ \\
\midrule
\rowcolor{blue!10}
\textbf{SparSTAR} & \textbf{25.56} & \textbf{0.793} & \textbf{0.123} & \textbf{83.48} & 1.30$\times$ \\
\midrule
\multicolumn{6}{c}{\textbf{480p Image-to-Video (I2V)}} \\
\addlinespace[2pt]
InfinityStar & -- & -- & -- & 80.15 & 1.00$\times$ \\
\midrule
\addlinespace[1pt]
FastSTAR & 23.86 & 0.750 & 0.234 & -- & 1.47$\times$ \\
\midrule
\rowcolor{blue!10}
\textbf{SparSTAR} & \textbf{26.14} & \textbf{0.813} & \textbf{0.111} & \textbf{79.03} & 1.30$\times$ \\
\bottomrule
\end{tabularx}
\endgroup
\caption{Quality and efficiency for 5-second, 81-frame 480p T2V and I2V
generation. All T2V evaluations use the \texttt{refined\_prompt} strings
distributed with InfinityStar. Dashes indicate non-applicable or unreported
metrics.}
\label{tab:supp-480p-results}
\end{table}
\begin{table*}[!t]
\centering
\begingroup
\small
\setlength{\tabcolsep}{3pt}
\renewcommand{\arraystretch}{1.15}
\begin{tabular*}{\textwidth}{@{\extracolsep{\fill}}l*{8}{c}@{}}
\toprule
\multicolumn{9}{c}{\textbf{720p T2V: 16 VBench Dimensions ($\uparrow$)}} \\
\midrule
\textbf{Method}
& \makecell{Subject\\Consistency}
& \makecell{Background\\Consistency}
& \makecell{Temporal\\Flickering}
& \makecell{Motion\\Smoothness}
& \makecell{Dynamic\\Degree}
& \makecell{Aesthetic\\Quality}
& \makecell{Imaging\\Quality}
& \makecell{Object\\Class} \\
\midrule
InfinityStar & 0.9513 & 0.9560 & 0.9845 & 0.9873 & 0.6833 & 0.6429 & 67.690 & 0.9741 \\
SparSTAR     & 0.9513 & 0.9567 & 0.9846 & 0.9873 & 0.6861 & 0.6422 & 67.684 & 0.9731 \\
\midrule
\textbf{Method}
& \makecell{Multiple\\Objects}
& \makecell{Human\\Action}
& Color
& \makecell{Spatial\\Relationship}
& Scene
& \makecell{Appearance\\Style}
& \makecell{Temporal\\Style}
& \makecell{Overall\\Consistency} \\
\midrule
InfinityStar & 0.8750 & 0.9780 & 0.8850 & 0.8380 & 0.5515 & 0.2082 & 0.2579 & 0.2766 \\
SparSTAR     & 0.8745 & 0.9780 & 0.8854 & 0.8437 & 0.5526 & 0.2082 & 0.2580 & 0.2767 \\
\bottomrule
\end{tabular*}

\begin{tabular*}{\textwidth}{@{\extracolsep{\fill}}l*{9}{c}@{}}
\toprule
\multicolumn{10}{c}{\textbf{720p I2V: 9 VBench-I2V Dimensions ($\uparrow$)}} \\
\midrule
\textbf{Method}
& \makecell{I2V\\Subject}
& \makecell{I2V\\Background}
& \makecell{Subject\\Consistency}
& \makecell{Background\\Consistency}
& \makecell{Temporal\\Flickering}
& \makecell{Motion\\Smoothness}
& \makecell{Aesthetic\\Quality}
& \makecell{Imaging\\Quality}
& \makecell{Dynamic\\Degree} \\
\midrule
InfinityStar & 0.9800 & 0.9842 & 0.9461 & 0.9580 & 0.9810 & 0.9902 & 0.5973 & 70.554 & 0.4507 \\
SparSTAR     & 0.9799 & 0.9841 & 0.9460 & 0.9577 & 0.9811 & 0.9902 & 0.5969 & 70.456 & 0.4451 \\
\bottomrule
\end{tabular*}
\endgroup
\caption{Fine-grained VBench results for 720p generation. T2V uses the 16
standard VBench dimensions. I2V uses two input-condition alignment
dimensions and seven video-quality dimensions.}
\label{tab:supp-vbench-720p}
\end{table*}
\begin{table*}[!t]
\centering
\begingroup
\small
\setlength{\tabcolsep}{3pt}
\renewcommand{\arraystretch}{1.15}
\begin{tabular*}{\textwidth}{@{\extracolsep{\fill}}l*{8}{c}@{}}
\toprule
\multicolumn{9}{c}{\textbf{480p T2V: 16 VBench Dimensions ($\uparrow$)}} \\
\midrule
\textbf{Method}
& \makecell{Subject\\Consistency}
& \makecell{Background\\Consistency}
& \makecell{Temporal\\Flickering}
& \makecell{Motion\\Smoothness}
& \makecell{Dynamic\\Degree}
& \makecell{Aesthetic\\Quality}
& \makecell{Imaging\\Quality}
& \makecell{Object\\Class} \\
\midrule
InfinityStar & 0.9416 & 0.9577 & 0.9812 & 0.9831 & 0.7361 & 0.6609 & 66.011 & 0.9786 \\
SparSTAR     & 0.9381 & 0.9545 & 0.9820 & 0.9823 & 0.7500 & 0.6543 & 64.298 & 0.9826 \\
\midrule
\textbf{Method}
& \makecell{Multiple\\Objects}
& \makecell{Human\\Action}
& Color
& \makecell{Spatial\\Relationship}
& Scene
& \makecell{Appearance\\Style}
& \makecell{Temporal\\Style}
& \makecell{Overall\\Consistency} \\
\midrule
InfinityStar & 0.8986 & 0.9800 & 0.8810 & 0.8255 & 0.5240 & 0.2231 & 0.2574 & 0.2786 \\
SparSTAR     & 0.8803 & 0.9800 & 0.8875 & 0.8280 & 0.5298 & 0.2219 & 0.2575 & 0.2780 \\
\bottomrule
\end{tabular*}

\begin{tabular*}{\textwidth}{@{\extracolsep{\fill}}l*{9}{c}@{}}
\toprule
\multicolumn{10}{c}{\textbf{480p I2V: 9 VBench-I2V Dimensions ($\uparrow$)}} \\
\midrule
\textbf{Method}
& \makecell{I2V\\Subject}
& \makecell{I2V\\Background}
& \makecell{Subject\\Consistency}
& \makecell{Background\\Consistency}
& \makecell{Temporal\\Flickering}
& \makecell{Motion\\Smoothness}
& \makecell{Aesthetic\\Quality}
& \makecell{Imaging\\Quality}
& \makecell{Dynamic\\Degree} \\
\midrule
InfinityStar & 0.9751 & 0.9812 & 0.9517 & 0.9592 & 0.9825 & 0.9909 & 0.5996 & 68.948 & 0.4085 \\
SparSTAR     & 0.9724 & 0.9792 & 0.9473 & 0.9561 & 0.9830 & 0.9904 & 0.5887 & 67.392 & 0.3577 \\
\bottomrule
\end{tabular*}
\endgroup
\caption{Fine-grained VBench results for 480p generation. T2V uses the 16
standard VBench dimensions. I2V uses two input-condition alignment
dimensions and seven video-quality dimensions.}
\label{tab:supp-vbench-480p}
\end{table*}

\subsection{Fine-Grained VBench Results}
\label{sec:supp-fine-grained-vbench}
Tables~\ref{tab:supp-vbench-720p} and~\ref{tab:supp-vbench-480p} organize the
16 VBench dimensions for T2V and the nine VBench-I2V dimensions for I2V
generation at 720p and 480p. The tables include rows for InfinityStar and
SparSTAR under the corresponding generation settings.

At 720p the two methods are hard to tell apart dimension by dimension. Subject
consistency is identical to four decimals (0.9513), motion smoothness is
identical as well, temporal flickering differs only in the fourth decimal, and
imaging quality moves by 0.006 points out of roughly 68. The largest single
change is dynamic degree, which rises slightly from 0.6833 to 0.6861, so
sparsification does not flatten motion at this resolution. The I2V half of the
table behaves the same way: the two input-condition alignment dimensions match
to within 0.0001.

At 480p the same dimensions separate a little further. Imaging quality falls
from 66.011 to 64.298 and multiple objects from 0.8986 to 0.8803, while dynamic
degree again moves upward, from 0.7361 to 0.7500. The clearest gap in either
table is on the 480p I2V side, where dynamic degree drops from 0.4085 to 0.3577.
Apart from that dimension and imaging quality, which falls by 1.56 on its
0--100 scale, every remaining I2V dimension stays within 0.011 of dense
InfinityStar.

\subsection{Per-Dimension Differences with Confidence Intervals}
\label{sec:supp-vbench-ci}
Table~\ref{tab:supp-vbench-ci} is a matched-clip robustness check from a
separate 720p T2V batch with prompt scopes different from the full-suite
protocol used for Main Table~\MainTabMain{}. It is not a confidence-interval
analysis of the headline means. Among clips scored by both methods, five small
differences have intervals excluding zero, whereas the dynamic-degree interval
straddles zero.

\begin{table*}[!t]
\centering
\begingroup
\small
\setlength{\tabcolsep}{6pt}
\renewcommand{\arraystretch}{1.15}
\begin{tabular*}{\textwidth}{@{\extracolsep{\fill}}lccccc@{}}
\toprule
\textbf{Dimension} & \textbf{Clips} ($n$) & \textbf{InfinityStar} & \textbf{SparSTAR}
& \textbf{Difference} & \textbf{95\% CI} \\
\midrule
Subject consistency & 4730 & 0.9553 & 0.9517 & $-0.0036$ & [$-0.0038$, $-0.0035$] \\
Background consistency & 430 & 0.9369 & 0.9353 & $-0.0016$ & [$-0.0025$, $-0.0008$] \\
Temporal flickering & 375 & 0.9832 & 0.9838 & $+0.0007$ & [$+0.0006$, $+0.0007$] \\
Motion smoothness & 360 & 0.9852 & 0.9863 & $+0.0011$ & [$+0.0010$, $+0.0012$] \\
Aesthetic quality & 465 & 0.6196 & 0.6182 & $-0.0014$ & [$-0.0026$, $-0.0003$] \\
Dynamic degree & 360 & 0.7361 & 0.7389 & $+0.0028$ & [$-0.0083$, $+0.0139$] \\
\bottomrule
\end{tabular*}
\endgroup
\caption{Matched-clip VBench differences with 95\% bootstrap intervals (10,000
resamples) from a separate 720p T2V batch. The retained score pools use prompt
scopes different from the full-suite protocol; this is therefore a robustness
check, not a CI analysis of the headline means. Imaging quality lacks
per-video records in this batch.}
\label{tab:supp-vbench-ci}
\end{table*}
\begin{table*}[!t]
\begingroup
\centering
\small
\setlength{\tabcolsep}{5pt}
\renewcommand{\arraystretch}{1.15}
\begin{tabularx}{\textwidth}{>{\raggedright\arraybackslash}p{0.30\textwidth} *{5}{>{\centering\arraybackslash}X}}
\toprule
\textbf{Selection Strategy}
& \textbf{PSNR} ($\uparrow$)
& \textbf{SSIM} ($\uparrow$)
& \textbf{LPIPS} ($\downarrow$)
& \textbf{VBench} ($\uparrow$)
& \textbf{Speedup} ($\uparrow$) \\
\midrule
Dense InfinityStar (reference)
& -- & -- & -- & 83.45 & 1.00$\times$ \\
\rowcolor{blue!10}
\textbf{Fresh aggregated-QK (ours)}
& \textbf{29.09} & \textbf{0.889} & \textbf{0.075}
& \textbf{83.13} & \textbf{2.43$\times$} \\
Sliding window
& 27.43 & 0.847 & 0.112 & 82.86 & 2.43$\times$ \\
Cross-scale reuse (SparVAR-style)
& 23.02 & 0.743 & 0.252 & 78.53 & 2.48$\times$ \\
Random blocks
& 19.37 & 0.616 & 0.423 & 69.79 & 2.43$\times$ \\
\bottomrule
\end{tabularx}
\endgroup
\caption{Quantitative comparison of selection strategies for the 720p T2V
analysis associated with Figure~\MainFigMassGap{} in the main paper. Reconstruction metrics
are measured against paired dense InfinityStar outputs.}
\label{tab:supp-selection-strategy}
\end{table*}
\section{Additional Ablation Study}
\label{sec:supp-additional-ablation}

\subsection{Selection Strategy Analysis}
\label{sec:supp-selection-strategy-analysis}
Table~\ref{tab:supp-selection-strategy} provides a quantitative comparison of
the methods analyzed in Figure~\MainFigMassGap{} of the main paper. Beyond
measuring retained attention mass, we run full video-generation inference and
evaluate the resulting videos. At a matched 2.43$\times$ speedup, Fresh
aggregated-QK (Ours) achieves the highest PSNR, SSIM, and VBench and the lowest
LPIPS among the evaluated sparse methods. The ordering is wide enough to read
without statistics: fresh selection is 1.66~dB above a sliding window, 6.07~dB
above cross-scale reuse, and 9.72~dB above random blocks, and all four arms run
at essentially the same speed. This shows that fresh aggregated-QK translates
the attention-mass advantage observed in Figure~\MainFigMassGap{} into better
generation fidelity. Evaluation definitions and prompt protocols are provided in
Section~\ref{sec:supp-dataset-evaluation-details}.

\begin{table}[!tb]
\begingroup
\centering
\small
\setlength{\tabcolsep}{1.5pt}
\renewcommand{\arraystretch}{1.10}
\begin{tabular*}{\linewidth}{@{\extracolsep{\fill}}ll*{4}{c}@{}}
\toprule
\makecell{\textbf{Block}\\\textbf{Size}}
& \textbf{Kernel}
& \makecell{\textbf{PSNR}\\($\uparrow$)}
& \makecell{\textbf{SSIM}\\($\uparrow$)}
& \makecell{\textbf{LPIPS}\\($\downarrow$)}
& \makecell{\textbf{Latency}\\(s)} \\
\midrule
32 & Pure tensor$^\dagger$
& 29.97 & 0.914 & 0.055 & 180.9 \\
64 & Pure tensor$^\dagger$ & 29.62 & 0.910 & 0.057 & 117.6 \\
\textbf{128 (ours)} & FlexAttention
& 29.09 & 0.889 & 0.075 & \textbf{27.9} \\
256 & FlexAttention & 28.69 & 0.879 & 0.085 & 43.7 \\
\bottomrule
\end{tabular*}
\endgroup
\caption{Block-size ablation in the InfinityStar environment with
PyTorch~2.5.1. $\dagger$ For block sizes below 128, no valid FlexAttention
kernel configuration was available in our tested setup; we therefore used a
forced pure-tensor fallback.}
\label{tab:supp-block-size-ablation}
\end{table}

\subsection{Block-Size Ablation}
\label{sec:supp-block-size-ablation}
Table~\ref{tab:supp-block-size-ablation} studies why we fix the deployed block
size to 128. We compare block sizes of 32, 64, 128, and 256 using the same
InfinityStar inference setting. In our PyTorch~2.5.1 environment,
FlexAttention does not provide a valid kernel configuration for block sizes
below 128. The 32 and 64 settings therefore require a forced pure-tensor
fallback and incur substantial execution overhead, with end-to-end latencies
of 180.9~s and 117.6~s, respectively.

Block size 128 reduces latency to 27.9~s while maintaining a PSNR of 29.09,
an SSIM of 0.889, and an LPIPS of 0.075. Increasing the block size to 256
degrades all three reconstruction metrics and raises latency to 43.7~s,
approximately 1.57$\times$ that of block size 128. Reading the table as a whole,
quality improves monotonically as blocks get smaller, but only by 1.28~dB from
256 down to 32, whereas latency swings by a factor of six over the same range.
Block size 128 sits at the point where the kernel is still available and the
remaining quality on offer no longer pays for its cost, which is why we adopt it
as the default deployment setting.
\section{Additional Qualitative Results}
\label{sec:supp-additional-qualitative-results}

\subsection{Qualitative Comparisons}
\label{sec:supp-qualitative-comparisons}
\begin{figure*}[p]
\centering
\includegraphics[width=\textwidth,height=0.90\textheight,keepaspectratio]{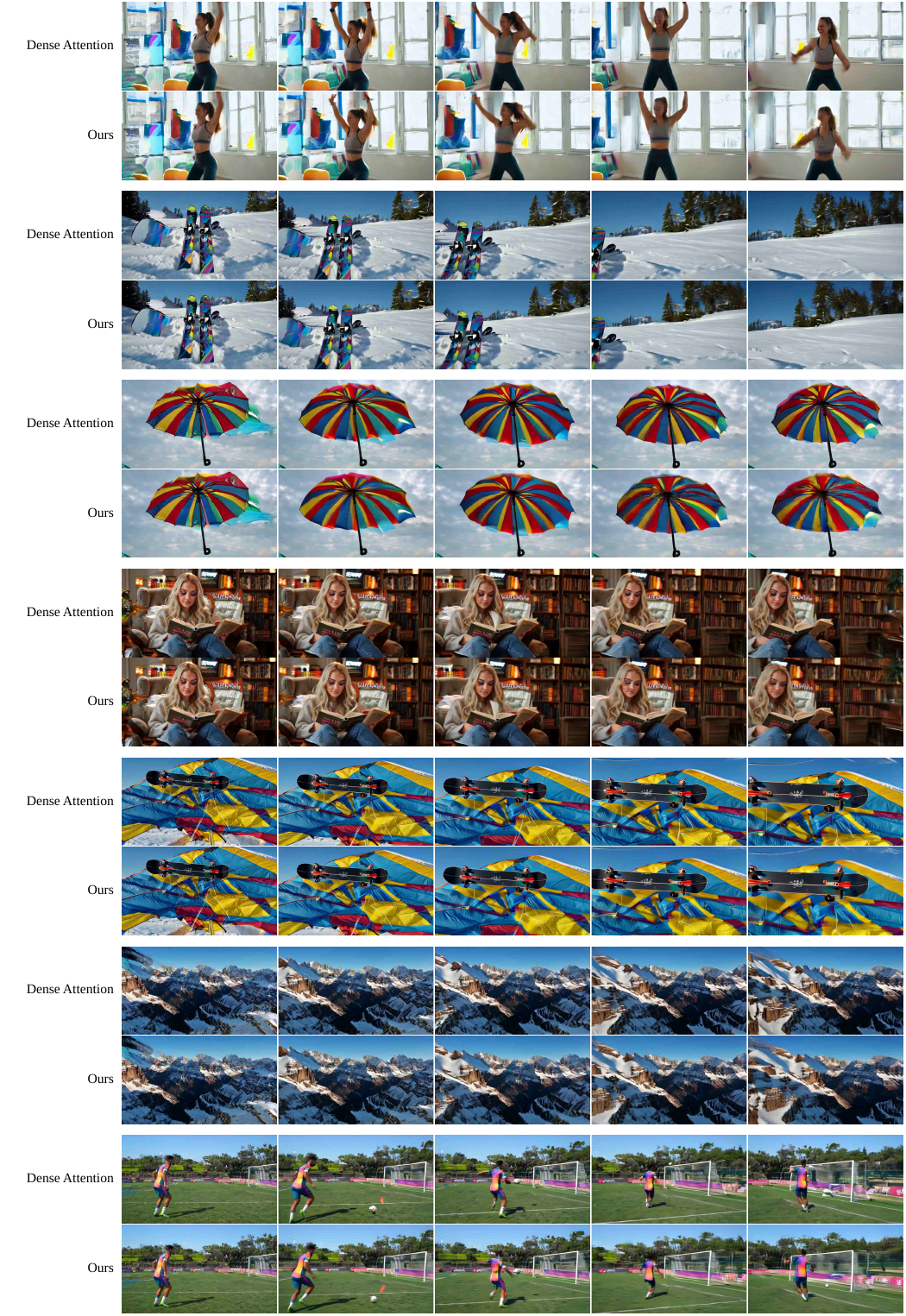}
\caption{Qualitative comparison between dense InfinityStar (top) and SparSTAR
(bottom) for 480p five-second T2V generation. SparSTAR preserves high
pixel-level fidelity and achieves visual quality comparable to dense
InfinityStar.}
\label{fig:supp-qual-t2v-480p-5s}
\end{figure*}

\begin{figure*}[p]
\centering
\includegraphics[width=\textwidth,height=0.90\textheight,keepaspectratio]{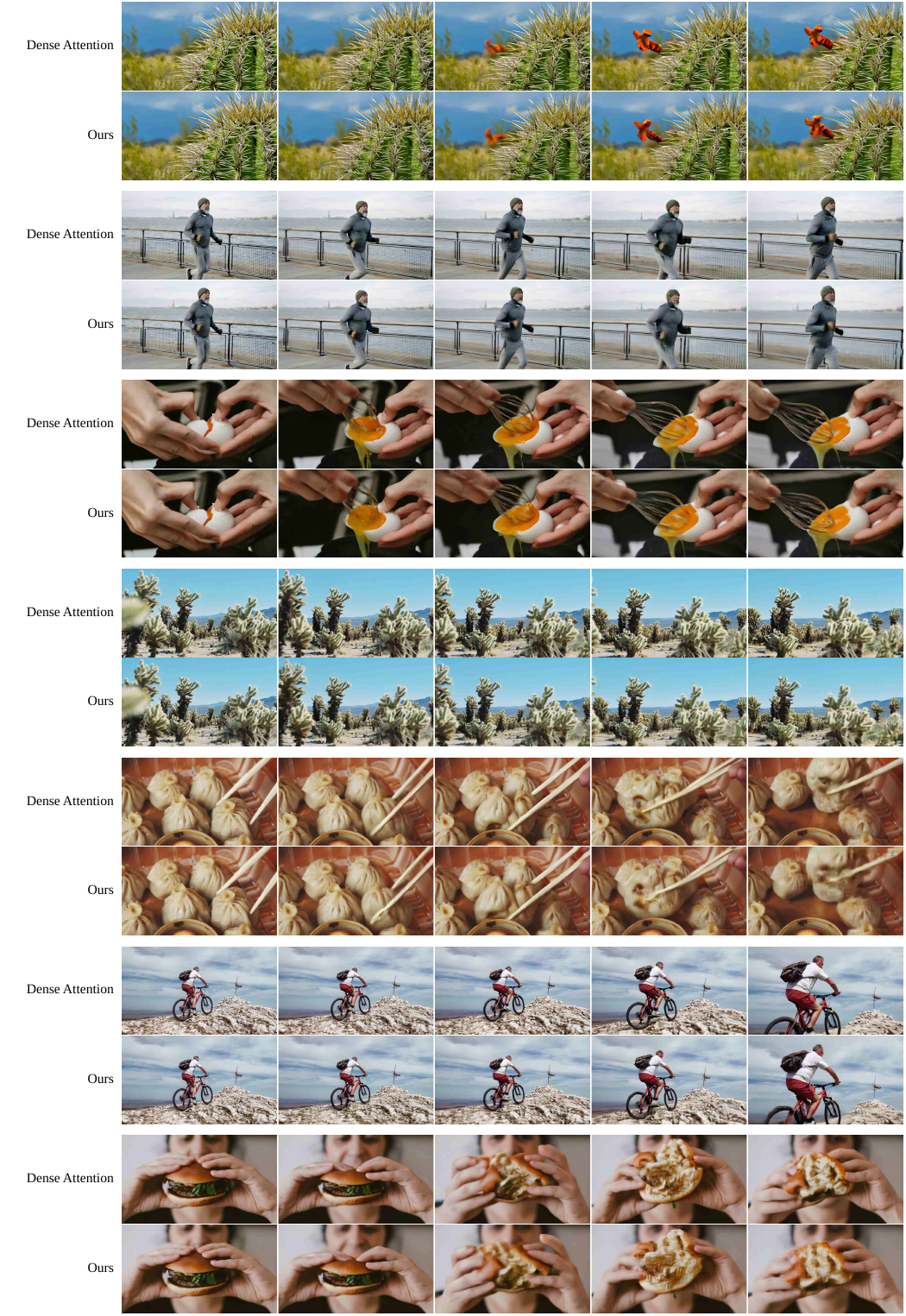}
\caption{Qualitative comparison between dense InfinityStar (top) and SparSTAR
(bottom) for 480p five-second I2V generation. SparSTAR preserves high
pixel-level fidelity and achieves visual quality comparable to dense
InfinityStar.}
\label{fig:supp-qual-i2v-480p-5s}
\end{figure*}

\begin{figure*}[p]
\centering
\includegraphics[width=\textwidth,height=0.90\textheight,keepaspectratio]{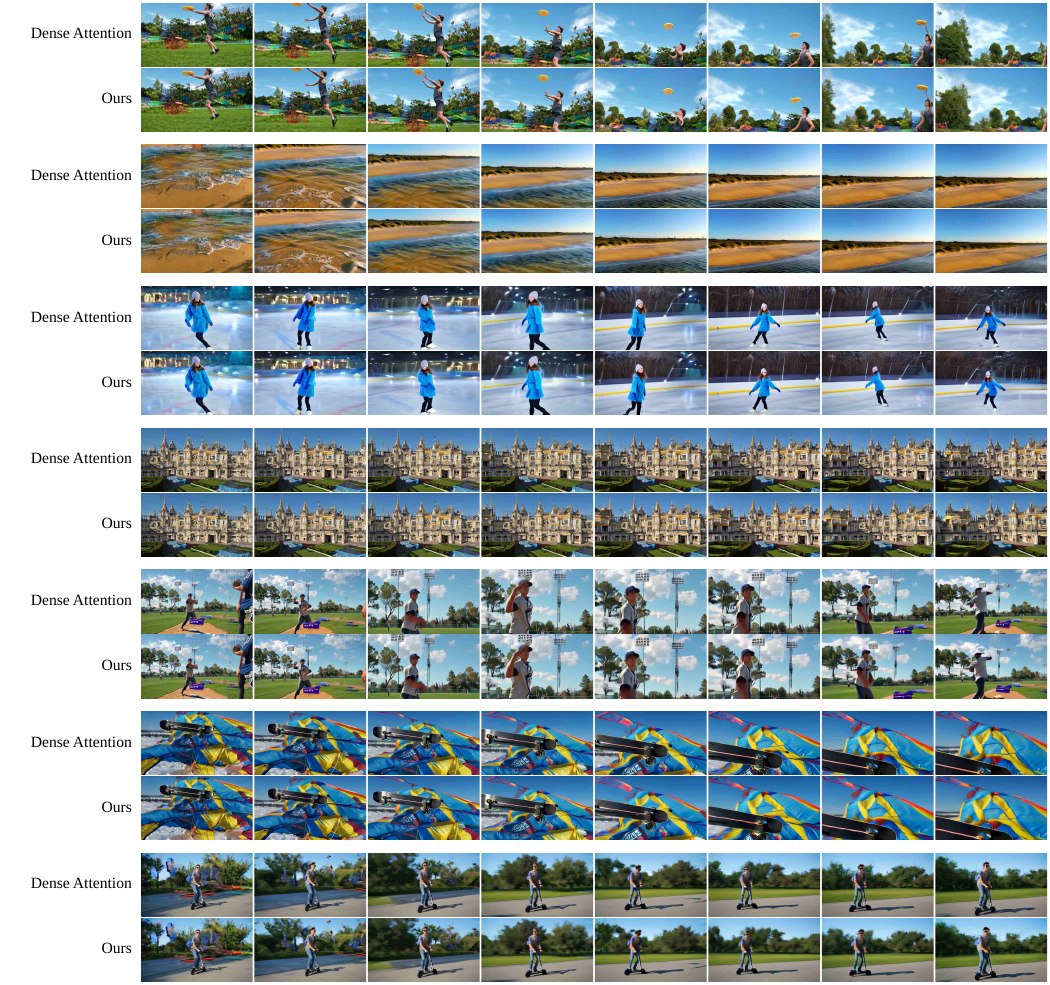}
\caption{Qualitative comparison between dense InfinityStar (top) and SparSTAR
(bottom) for 480p ten-second T2V generation. SparSTAR preserves high
pixel-level fidelity and achieves visual quality comparable to dense
InfinityStar.}
\label{fig:supp-qual-t2v-480p-10s}
\end{figure*}

\begin{figure*}[p]
\centering
\includegraphics[width=\textwidth,height=0.90\textheight,keepaspectratio]{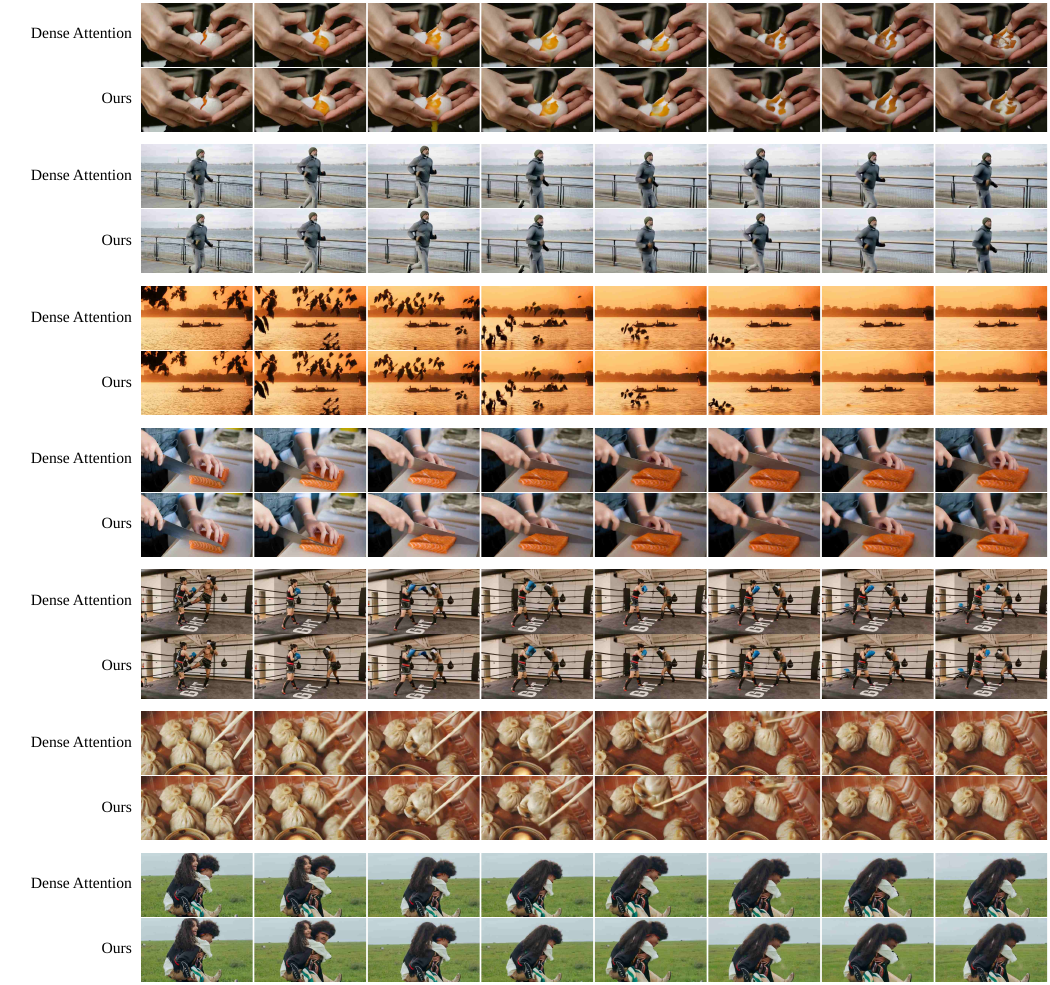}
\caption{Qualitative comparison between dense InfinityStar (top) and SparSTAR
(bottom) for 480p ten-second I2V generation. SparSTAR preserves high
pixel-level fidelity and achieves visual quality comparable to dense
InfinityStar.}
\label{fig:supp-qual-i2v-480p-10s}
\end{figure*}

\begin{figure*}[p]
\centering
\includegraphics[width=\textwidth,height=0.90\textheight,keepaspectratio]{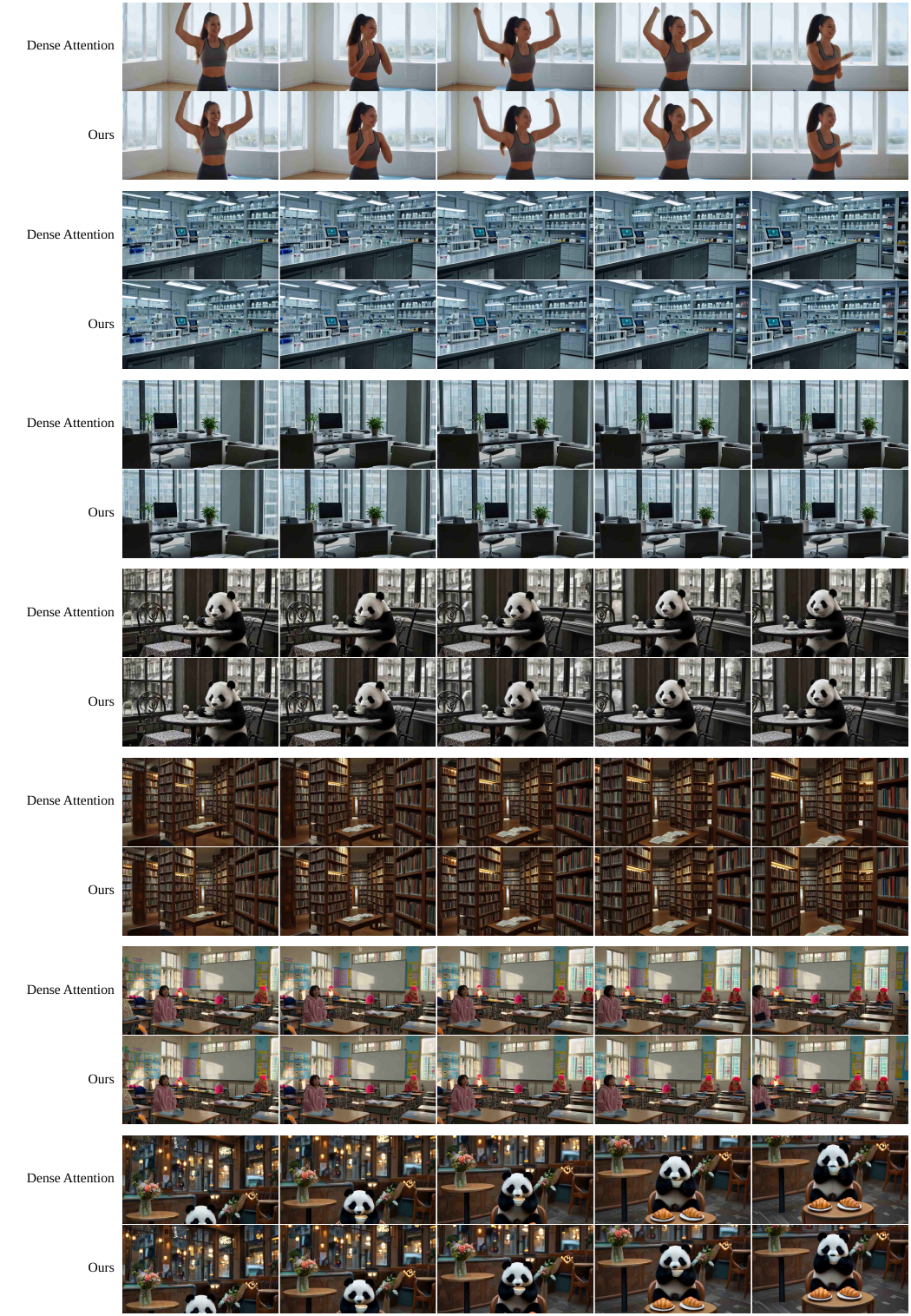}
\caption{Qualitative comparison between dense InfinityStar (top) and SparSTAR
(bottom) for 720p five-second T2V generation. SparSTAR preserves high
pixel-level fidelity and achieves visual quality comparable to dense
InfinityStar.}
\label{fig:supp-qual-t2v-720p-5s}
\end{figure*}

\begin{figure*}[p]
\centering
\includegraphics[width=\textwidth,height=0.90\textheight,keepaspectratio]{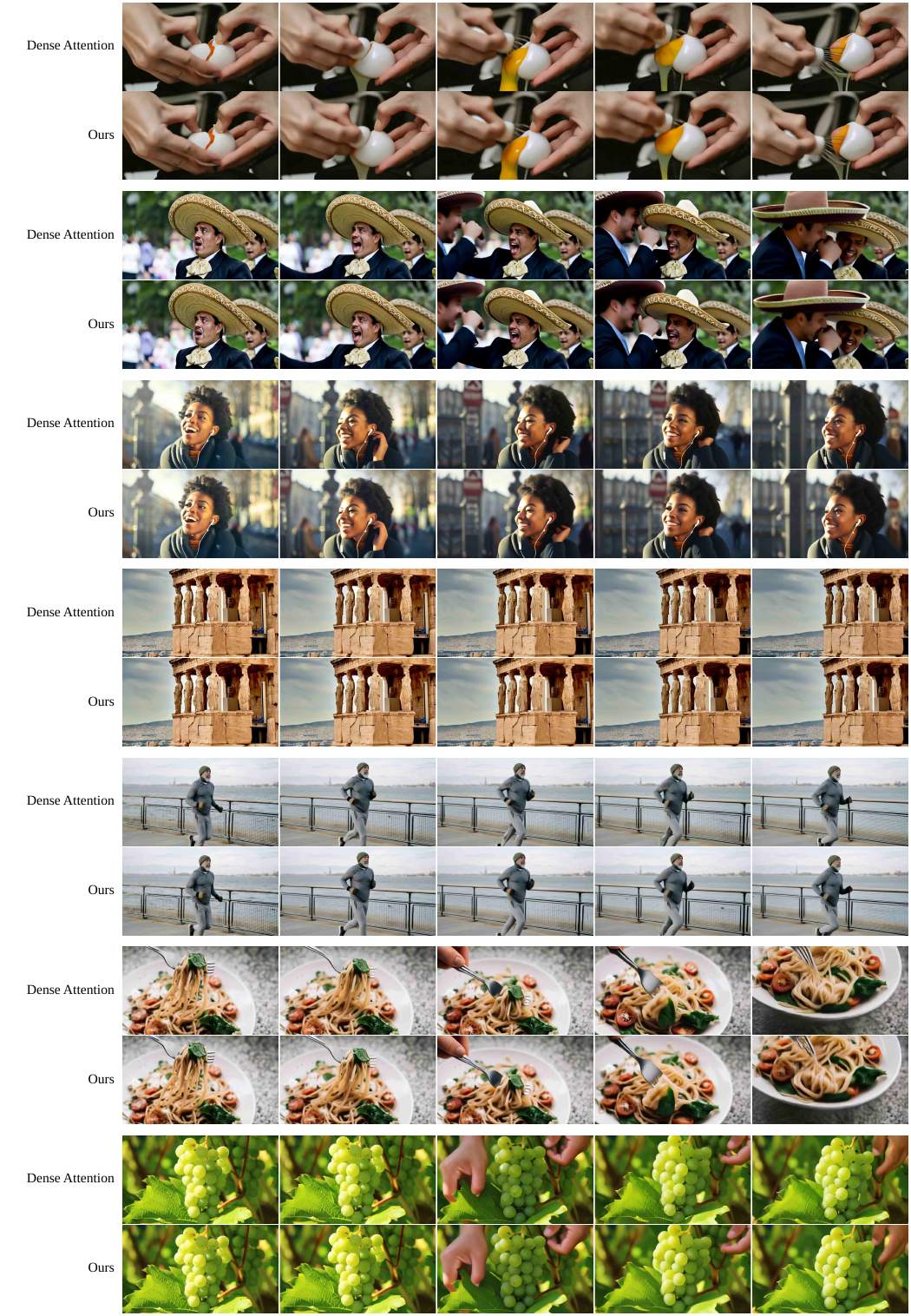}
\caption{Qualitative comparison between dense InfinityStar (top) and SparSTAR
(bottom) for 720p five-second I2V generation. SparSTAR preserves high
pixel-level fidelity and achieves visual quality comparable to dense
InfinityStar.}
\label{fig:supp-qual-i2v-720p-5s}
\end{figure*}

Figures~\ref{fig:supp-qual-t2v-480p-5s}--\ref{fig:supp-qual-i2v-720p-5s}
compare dense InfinityStar with SparSTAR for T2V and I2V generation at 480p and
720p. Across resolutions, durations, and tasks, SparSTAR preserves pixel-level
fidelity and matches the visual quality of the dense reference.

\paragraph{Comparison protocol.}
Each dense--sparse pair uses the same prompt, seed, resolution, duration, and,
for I2V, conditioning image. Frames are shown at matched temporal indices.
The five-second examples assess within-clip subject, background, and texture
consistency, while the ten-second examples additionally reveal behavior across
the Clip~1--Clip~2 boundary. The I2V comparisons also show how closely each
method preserves the subject and appearance of the conditioning image.

\FloatBarrier

\end{document}